%% file: main.tex
\documentclass[10pt,twocolumn,letterpaper]{article}

\usepackage[pagenumbers]{cvpr}%
\usepackage{multicol}
\usepackage{algorithm}
\usepackage{soul}
\usepackage{xspace}\xspace
\usepackage{adjustbox}
\usepackage{array}
\usepackage{algpseudocode}
\usepackage{amsmath}
\usepackage{dblfloatfix}
\usepackage{transparent}
\usepackage{multirow} 
\usepackage{algorithm}
\usepackage{listings}

\usepackage{etoolbox}
\input{preamble}

\definecolor{cvprblue}{rgb}{0.21,0.49,0.74}
\usepackage{hyperref}
\hypersetup{pagebackref,breaklinks,colorlinks,allcolors=cvprblue}
\PassOptionsToPackage{capitalize}{cleveref}
\usepackage{cleveref}

\crefname{algorithm}{Alg.}{Algs.}
\Crefname{algorithm}{Alg.}{Algs.}
\crefname{figure}{Fig.}{Figs.} 
\Crefname{figure}{Fig.}{Figs.} 

\def\paperID{12603} %
\def\confName{CVPR}
\def\confYear{2025}

\title{SegAgent: Exploring Pixel Understanding Capabilities in MLLMs by Imitating Human Annotator Trajectories}

\author{
Muzhi Zhu$^{1,2}$~~
Yuzhuo Tian$^1$~~
Hao Chen$^1$\thanks{Correspondence should be addressed to HC, QG, CS}~~
Chunluan Zhou$^2$~~ \\
Qingpei Guo$^2$\footnotemark[1]~~
Yang Liu$^1$~~
Ming Yang$^2$~~
Chunhua Shen$^1$\footnotemark[1]
\\[.25cm]
$ ^1$ Zhejiang University, China 
~~~~~
$ ^2$ 
Ant Group.
}

\begin{document}
\maketitle
\begin{abstract}

While MLLMs have demonstrated adequate image understanding capabilities, they still struggle with pixel-level comprehension, limiting their practical applications. Current evaluation tasks like VQA and visual grounding remain too coarse to assess fine-grained pixel comprehension accurately. Though segmentation is foundational for pixel-level understanding, existing methods often require MLLMs to generate implicit tokens, decoded through external pixel decoders. This approach disrupts the MLLM’s text output space, potentially compromising language capabilities and reducing flexibility and extensibility, while failing to reflect the model’s intrinsic pixel-level understanding.
Thus, We introduce the Human-Like Mask Annotation Task (HLMAT), a new paradigm where MLLMs mimic human annotators using interactive segmentation tools. Modeling segmentation as a multi-step Markov Decision Process, HLMAT enables MLLMs to iteratively generate text-based click points, achieving high-quality masks without architectural changes or implicit tokens. Through this setup, we develop SegAgent, a model fine-tuned on human-like annotation trajectories, which achieves performance comparable to SOTA methods and supports additional tasks like mask refinement and annotation filtering.
HLMAT provides a protocol for assessing fine-grained pixel understanding in MLLMs and introduces a vision-centric, multi-step decision-making task that facilitates exploration of MLLMs’ visual reasoning abilities. Our adaptations of policy improvement method StaR and PRM guided tree search further enhance model robustness in complex segmentation tasks, laying a foundation for future advancements in fine-grained visual perception and multi-step decision-making for MLLMs. Code will be released at \url{https://github.com/aim-uofa/SegAgent}.
\end{abstract} 
\section{Introduction}

Recent advancements~\cite{OpenAI2023GPT4,Gemini} in large language models (LLMs) have brought them to near-human levels in many domains. However, multimodal LLMs (MLLMs) still face significant limitations in visual tasks \cite{tong2024cambrian,tong2024eyes}, especially in complex pixel-level understanding \cite{ying2024mmt}, which restricts their applicability in real-world scenarios. Despite efforts to develop MLLM vision capabilities through tasks like Visual Question Answering (VQA)\cite{blip,blip2}, and visual grounding \cite{bai2023qwen,wang2024qwen2}, these tasks fail to assess fine-grained pixel-level comprehension effectively.
On the other hand, tasks like UI Operations~\cite{zhang2023you,zhang2024android} and Robotic Control~\cite{shridhar2020alfred} require pixel-level precision but are often constrained to specialized environments with limited, heterogeneous data, making them difficult to scale to complex, open-world scenarios.

Segmentation \cite{liu2023matcher,zhu2023segprompt,zhu2024unleashing,fan2024divergen,zhu2024generative,liu2025simple,zhao2025diception} is one of the most critical visual tasks for reflecting pixel-level understanding capabilities. Moreover, there is a wealth of segmentation data \cite{sam} available, which can sufficiently meet the needs of complex open-world scenarios. Establishing a method that enables MLLMs to effectively leverage this large-scale data is essential to further explore and enhance their pixel-level understanding capabilities. However, current approaches \cite{lisa,lisa++,rasheed2024glamm,xia2023gsva,zhang2025psalm} integrating MLLMs with segmentation tasks often rely on implicit token outputs and require additional pixel decoders. While effective, these methods alter the original MLLM output space, potentially compromising semantic generalization by shifting away from language-based outputs. The use of implicit tokens also limits extensibility, complicating efforts to implement a unified textual framework across diverse tasks. Furthermore, reliance on an extra pixel decoder makes it challenging to accurately reflect the MLLM’s intrinsic pixel understanding capabilities \cite{siam2025pixfoundation, cao2024emerging}. There are also methods \cite{chen2022unified,liu2023polyformer,pramanick2024jack, lan2024text4segreimaginingimagesegmentation} that attempt to perform segmentation using purely textual outputs, such as polygon vertex sequences or sampled points along the mask contour.  However, this strategy limits mask precision, and generating lengthy coordinate sequences is particularly challenging. Even for a skilled human annotator, tracing the mask contour point by point is time-intensive and laborious \cite{lin2014microsoft}. 
Recent annotation practices \cite{sam} have introduced interactive segmentation tools \cite{sam,liu2023simpleclick} that reduce workload by allowing iterative mask refinement through positive and negative clicks.
It’s natural to ask whether current MLLMs can perform segmentation by emulating the trajectories of human annotators using interactive segmentation tools. This approach would enable the model to complete the segmentation task more easily in a pure-text format. 

Therefore, we propose a new segmentation paradigm, enabling MLLMs to imitate human annotators using interactive segmentation tools by modeling the segmentation task as a multi-step Markov Decision Process. We refer to this as the Human-Like Mask Annotation Task (HLMAT). 
Compared to previous vision tasks, 
HLMAT imposes greater demands on the model’s pixel-level understanding for both input and output.
This paradigm allows MLLMs to iteratively generate text-based coordinates without relying on specialized architectures or implicit tokens, supporting fair comparisons across different MLLMs. 
We believe HLMAT offers a novel evaluation protocol to explore and advance MLLMs’ fine-grained visual capabilities.

We first develop an automated algorithm to transform existing datasets into annotation trajectories. Using the trajectories,  we fine-tune MLLMs, resulting in our model, SegAgent. Our experiments show that, after fine-tuning, various MLLMs achieve performance comparable to state-of-the-art methods. Additionally, we demonstrate that SegAgent not only excels at segmentation tasks but also supports mask refinement and annotation filtering effectively.

Furthermore, HLMAT, as a multi-step decision-making task, directly assesses pixel-level visual skills, unlike prior MLLM decision-making tasks that focus primarily on textual reasoning \cite{zhai2024fine} or simple UI-based operations \cite{bai2024digirl}. Our study lays the foundation for exploring MLLMs as vision-centered, multi-step decision-making agents, addressing the gap in pixel-level visual understanding in multi-step contexts.  In this work, we first examine commonly used referring segmentation datasets like RefCOCO \cite{refcoco} and find that their annotation complexity and quality are insufficient for multi-step decision-making. To support deeper exploration, we introduce high-quality reference segmentation datasets.

Next, we investigate whether certain enhancement methods for text-based LLM decision-making can be adapted to our proposed HLMAT task. We select StaR \cite{zelikman2022star}, a straightforward yet representative RL-style policy improvement method, and modify it for our purposes. Our findings reveal that incorporating the StaR algorithm, as opposed to solely relying on SFT data, significantly improves the model’s performance on the test dataset. Additionally, inspired by frameworks such as Tree-of-Thoughts \cite{yao2024tree}, we explore inference-stage techniques.
 By combining process reward modeling (PRM) \cite{lightman2023let, ma2023let} with tree search, we demonstrate that for complex scene segmentation tasks, implementing tree search effectively mitigates errors arising from inaccurate predictions, substantially enhancing the model's robustness in challenging segmentation scenarios. This work lays a strong foundation for advancing vision-centered, multi-step decision-making agents in future research.

Our main contributions are as follows:

\begin{itemize}
    \item  We propose Human-Like Mask Annotation Task (HLMAT), a new segmentation framework where MLLMs imitate human annotators using interactive tools, modeled as a multi-step Markov Decision Process. HLMAT raises pixel understanding demands and only requires text-based coordinate outputs, allowing fair comparison across MLLMs and serving as a new protocol for fine-grained visual capability assessment.

    \item  Using an automated algorithm to generate human-like annotation trajectories from existing datasets, we fine-tune MLLMs to create SegAgent. Experiments show SegAgent achieves competitive performance, supporting segmentation, mask refinement, and annotation filtering.

    \item  HLMAT lays a foundation for multi-step, vision-centered decision-making research. We introduce the HRES dataset, suitable for complex reasoning, and demonstrate that text-based decision improvement methods like StaR+ (training) and tree search with PRM (inference) effectively enhance segmentation performance.
\end{itemize}

\section{Related Work}
\label{sec:related_work}
\subsection{Pixel Understanding in MLLMs}

Many studies have explored pixel understanding in MLLMs by combining these models with segmentation tasks. However, most approaches  ~\cite{rasheed2024glamm,yuan2024osprey,lai2024lisa,ren2024pixellm,zhang2025psalm,zhang2024omg} require MLLMs to learn an implicit token and involve additional fine-tuning with a separate pixel decoder. Among them, LISA~\cite{lai2024lisa} was the first to integrate MLLMs with SAM~\cite{sam}, while subsequent works~\cite{rasheed2024glamm,zhang2025psalm}  introduced additional tasks and larger training datasets, so we did not include them in our experimental comparisons. Other methods attempt to let MLLMs perform segmentation through text-based outputs, such as using polygon vertex sequences \cite{pramanick2024jack} or randomly sampled points \cite{chen2025sam4mllm} where the model predicts if a point lies within the mask. However, this approach inherently limits mask precision.
Our work, on the other hand, achieves segmentation by imitating the trajectory of human annotators using interactive segmentation tools. This approach enables the model to perform highly detailed segmentation tasks through simple text-based outputs.

\subsection{MLLMs as Decision-Making Agents}

In the text domain, extensive research has explored how to adopt LLMs as decision-making agents and enhance their reasoning and decision-making capabilities. Some works~\citep{wei2022chain,yao2023react,yao2023tree} improve LLM reasoning through prompt engineering during inference, while others \cite{zelikman2022star,zhu2023fine} investigate policy improvement methods during training. However, research on decision-making agents in the visual domain remains limited. A few studies integrate MLLMs~\citep{mu2023embodiedgpt,chen2024vision} into training processes, but they train simple MLP or transformer layers to align action spaces without using text-based actions.

As RL4VLM \cite{zhai2024fine} emphasized, prior work primarily examined MLLM capabilities in non-interactive tasks. 
RL4VLM represents the first systematic approach to fine-tuning an entire VLM as a decision-making agent using reinforcement learning, directly interacting with the environment through open-ended text. Our work shares this advantage, offering a novel perspective on evaluating MLLM decision-making in interactive tasks. However, unlike RL4VLM, which focuses on poker-like tasks that are primarily text-based, our proposed task directly involves fine-grained visual perception, laying the groundwork for further exploration of vision-centered, multi-step decision-making agents. 

More recently, the success of DeepSeek-R1 \cite{guo2025deepseek} has further advanced RL training in the MLLM community. Since then, a series of related works \cite{chen2025r1v,yang2025r1one,ellm2025openr1,liu2025segzero} have emerged. However, these studies remain predominantly focused on text-centric reasoning. In contrast, our work took an earlier step in exploring vision-centric multi-step decision-making tasks, extending RL beyond high-level text interactions to fine-grained visual perception and action planning.

\subsection{Interactive Segmentation Tools}

The objective of interactive segmentation \cite{ding2020phraseclick,chen2022focalclick,sofiiuk2022reviving,liu2023simpleclick,huang2023interformer} is to segment regions of interest based on user-provided inputs, such as clicks, boxes, and scribbles. RITM \cite{sofiiuk2022reviving} leverages COCO \cite{caesar2018coco} and LVIS \cite{lvis} as training data, achieving substantial segmentation quality. SimpleClick \cite{liu2023simpleclick} was the first to explore the application of plain Vision Transformers (ViT) in interactive segmentation. InterFormer \cite{huang2023interformer} decouples feature encoding and prompt fusion, effectively mitigating the high-latency issue observed in SimpleClick. SAM \cite{sam} adopts a similar design, trained on the large-scale SA1B dataset, and achieves notable generalization capabilities, supporting box inputs.
However, existing interactive segmentation models, while capable of generating accurate masks, require human input. In contrast, our work aims to teach MLLMs to perform segmentation by imitating human annotators using interactive segmentation tools.

\section{Preliminaries and Task Formulation}
\label{sec:preliminaries}

Given an image $I$ and a text prompt $P$, a professional human annotator is able to accurately draw the mask $M$ corresponding to the object indicated by the text prompt $P$. However, drawing the mask $M$ directly is time-intensive. 
Modern annotators commonly utilize an interactive segmentation tool, iteratively providing a series of positive and negative click points based on the current image $I$ and mask $M$, making it easier to create a high-quality mask. This entire process can be modeled as a Markov Decision Process (MDP) consisting of \((\mathcal{S}, \mathcal{A}, T, R, \gamma)\), where:
\begin{itemize}

    \item \textbf{State} $s_t \in \mathcal{S}$: Represents the current status of the mask $M_t$ in the segmentation process, along with the history of all previous actions $a_{0:t-1}$, i.e., $s_t = (M_t, a_{0:t-1})$. The initial state $s_0$ corresponds to an empty mask, while the goal state $s_{\text{goal}}$ corresponds to the optimal target mask $M_{\text{target}}$.

    \item \textbf{Action} $a_t \in \mathcal{A}$: Here, each action $a_t$ refers to an annotator's click operation, consisting of two components: an attribute $\alpha_t$ and a coordinate $\mathbf{c}_t$. The attribute $\alpha_t \in \{+1, -1\}$ indicates whether the point is a positive or negative click, while the coordinate $\mathbf{c}_t \in [0, 1]^2$ represents the relative position of the point within the image, scaled to the $[0, 1]$ range for both $x$ and $y$ axes.

    \item \textbf{Transition Function} $T: \mathcal{S} \times \mathcal{A} \rightarrow \mathcal{S}$: Represents the environment's response to the current state-action pair $(s_t, a_t)$, mapping it to the next state $s_{t+1}$. In this task, the environment’s response is determined by the input image $I$ and the interactive segmentation network $F_{iter}$. Specifically, the next mask $M_{t+1}$ is generated by $F_{iter}(I, s_t, a_t)$, with $s_{t+1} = (M_{t+1}, a_{0:t})$.

    \item \textbf{Reward Function} $R: \mathcal{S} \rightarrow \mathbb{R}$: The reward function evaluates the accuracy of the current mask $M_t$ relative to the target mask $M_{\text{target}}$ using the Intersection over Union (IoU) metric. Specifically, $R(s_t) = \text{IoU}(M_t, M_{\text{target}})$, where a higher IoU score indicates a better match between $M_t$ and $M_{\text{target}}$.

\end{itemize}

Since we are focused solely on the quality of the final mask, we omit the discount factor $\gamma$ in our setup. 
Given that the action space can be fully represented by text, and the state space can be fully represented by images, a competent MLLM should be able to perform this task as effectively as a human annotator.
Our objective is to train a policy $\pi_\theta(a_t | s_t, I, P)$ for the MLLM that, given an image $I$ and a text prompt $P$, outputs an action $a_t$ at each time step $t$. The goal is to maximize the expected reward $R(s_T)$ within a limited number of steps $T$. We refer to this as the Human-Like Mask Annotation Task (HLMAT).
In practice, to enable the MLLM to support more flexible use cases—such as mask refinement—we remove the dependency on $a_{0:t-1}$ in $s_t$, allowing the model to determine the next action based solely on the current mask $M_t$. 

\section{SegAgent}

As mentioned in \cref{sec:preliminaries}, our goal is to train an MLLM to learn a policy $\pi_\theta(a_t | s_t, I, P)$ that imitates the annotator's trajectory. 
In practice, to enable the MLLM to support more flexible use cases—such as mask refinement—we remove the dependency on $a_{0:t-1}$ in $s_t$, allowing the model to determine the next action based solely on the current mask $M_t$, so the policy becomes $\pi_\theta(a_t | M_t, I, P)$.

\begin{figure}[t]
    \centering
    \vspace*{-0.2cm}
    \includegraphics[width=1\linewidth]{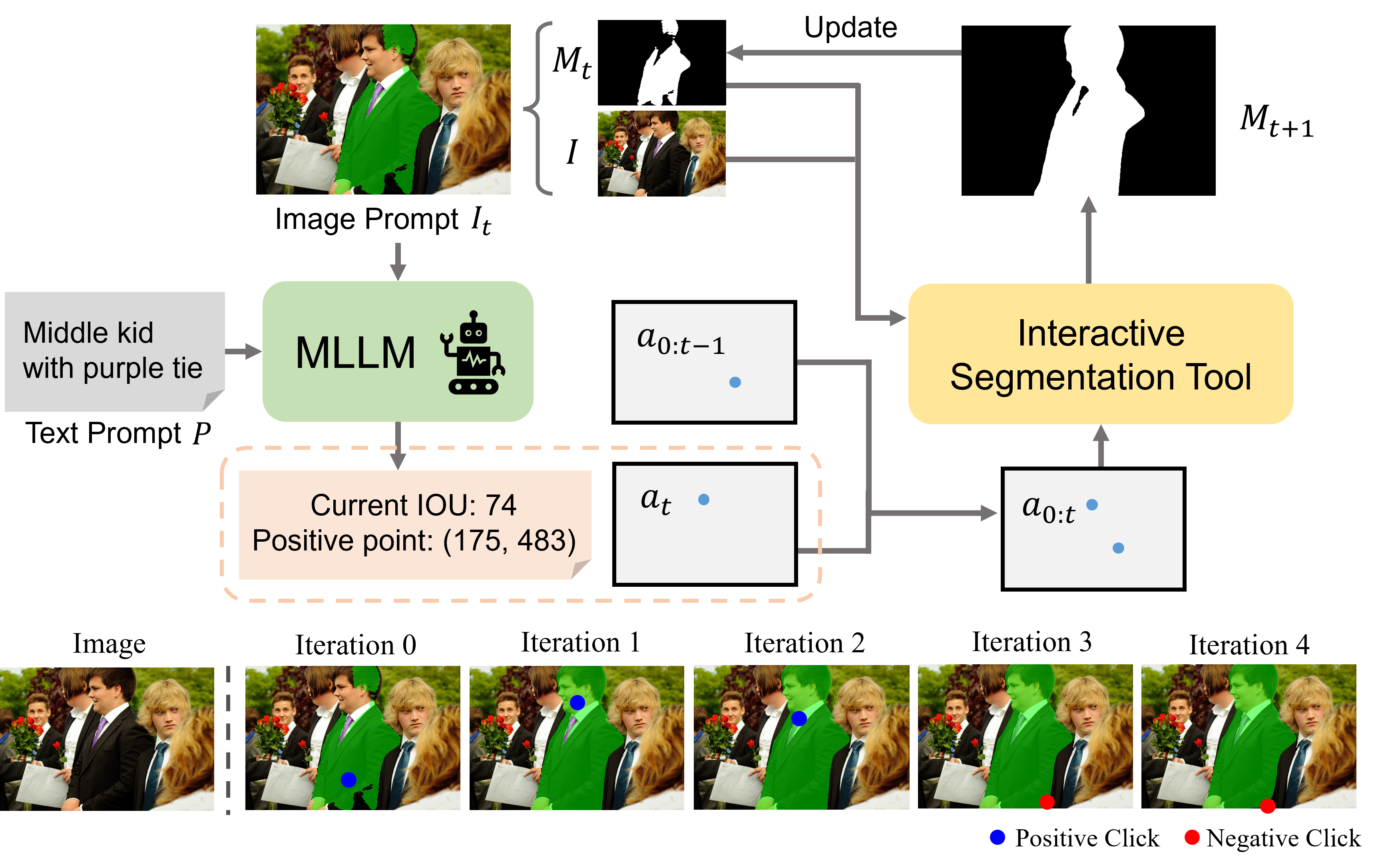}
    \vspace*{-0.6cm}
    \caption{The overall framework of SegAgent. The image below shows a complete set of trajectories. We visualize current action $a_t$ and the resulting mask 
    $M_{t+1}$ in one image.}
    \vspace*{-0.5cm}
    \label{fig:frameworks}
\end{figure}

As illustrated in \cref{fig:frameworks}, the current mask $M_t$ and the original image $I$ are processed as visual inputs to the MLLM. Following common visualization techniques used by human annotators, we overlay the current segmentation result $M_t$ as a semi-transparent green mask on the original image $I$ to create a new composite image $I_t$. This image $I_t$ serves as the final input for the MLLM, allowing the model to simultaneously perceive information about  $M_t$ and  $I$ in a single frame.

The text input $P$ to the MLLM includes two key components: (1) a description of the task and action space, and (2) a description of the target object to be segmented. This prompt design provides the model with context about the overall objective and the segmentation target, ensuring it can interpret both the visual input and the segmentation goal. Detailed prompt design can be found in the appendix.

After processing and analyzing the inputs, the MLLM outputs the next action $a_{t+1}$ in text format, such as "Positive point: (175,483)." 
The action is then converted into a format compatible with the interactive segmentation model via a predefined post-processing function. Subsequently, all previous actions $a_{0:t}$, the current mask $M_t$, and the original image $I$ are fed into the interactive segmentation model $F_{iter}$, which generates the updated mask $M_{t+1}$. 
This iterative process continues until the segmentation achieves a satisfactory result.

The remainder of this section is organized as follows: First, we describe our approach for collecting human-like annotation trajectories (\cref{subsec:trajectories}). Next, we analyze the two essential capabilities that a robust SegAgent should possess (\cref{subsec:what}). Inspired by traditional RL-style policy improvement methods, we then adapt the StaR method to our task (\cref{subsec:policy}). Finally, we combine the process reward model with tree search to enhance the model's robustness in complex scenarios (\cref{subsec:PRM}).

\subsection{Human Annotator Trajectories Generation}
\label{subsec:trajectories}
Suppose we have an annotated dataset $D_{seg} = \{(I, M_{target}, P)\}$, where each sample includes an image $I$, a target mask $M_{target}$, and a text prompt $P$. 
Ideally, recording all state-action pairs $(s_t, a_t)$ in real-time during the human annotation process would yield the desired trajectory $[s_0,a_0,s_1,a_1, ...,s_T,a_T]$. However, existing datasets lack such trajectories, and re-hiring human annotators to annotate the data while recording trajectories would be very costly. 
Thus, we consider whether it is possible to derive the trajectory dataset $D_{traj} = \{(I, M_{target}, P, [s_0,a_0,s_1,a_1, ...,s_T,a_T])\}$ from the existing dataset $D_{seg} = \{(I, M_{target}, P)\}$ using a rule-based method.
Thanks to researchers in interactive segmentation, an iterative click simulation strategy has been developed to automate the evaluation of interactive segmentation models. 
This strategy can be understood as a function $F_{sim}$, which takes as input the current mask $M_t$ and the ground truth mask $M_{target}$, and outputs the next action $a_{t+1}$, i.e., $a_{t+1} = F_{sim}(M_t, M_{target})$. 
Specifically, this function computes the false positive and false negative regions between the current mask $M_t$ and the ground truth mask $M_{target}$, placing the next click action at the center of the error region based on the size and position of these regions.
With the help of function $F_{sim}$ (\cref{alg:code}), we can simulate trajectories $[s_0, a_0, s_1, a_1, ..., s_T, a_T]$ based on $M_{target}$. However, the annotation quality of $M_{target}$ may vary, potentially generating suboptimal trajectories. This issue is especially pronounced toward the end of a trajectory (see \cref{fig:example}), where noise in $M_{target}$ can cause erroneous actions that misguide the MLLM during training. To address this, we introduce the following steps during generation:

\begin{figure}
    \centering
    \includegraphics[width=1\linewidth]{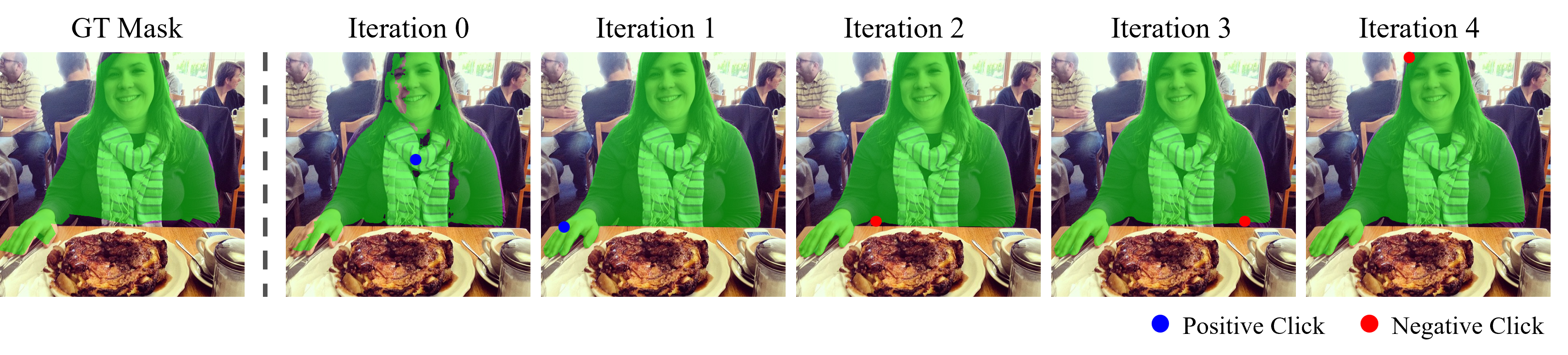}
    \vspace*{-0.8cm}
    \caption{An example of generated trajectory. We visualize current action $a_t$ and the resulting mask 
    $M_{t+1}$ in one image. Due to the noise from GT Mask, the action for Iteration 3,4 is meaningless}
    \vspace*{-0.6cm}
    \label{fig:example}
\end{figure}

\begin{itemize}
    \item \textbf{Limit Maximum Trajectory Length}: We set a maximum trajectory length $T$. 
    For lower-quality datasets, we reduce $T$ to avoid excessive noisy actions at the trajectory's end.
    \item \textbf{Terminate on Satisfactory Reward}: We calculate the reward $R(s_t)$ for the current state $s_t$, and if $R(s_t)$ reaches a threshold $\tau_{stop}$, we terminate the trajectory generation.
    \item \textbf{Discard Low-Impact Actions}: We evaluate the reward difference before and after executing an action $a_t$. If the reward difference $R(s_{t+1}) - R(s_t) < \tau_{diff}$, we discard action $a_t$. This is because abnormal actions can cause the interactive segmentation model to collapse, significantly reducing the quality of the resulting mask $M_{t+1}$.
\end{itemize}

To accommodate segmentation datasets of varying quality, we can adjust these three parameters to reduce noise in the generated trajectories. Following this process, we can generate the annotation trajectory dataset $D_{traj}$ from the existing segmentation dataset $D_{seg}$ and use these data to fine-tune the MLLM through instruction tuning, resulting in our SegAgent.
\subsection{What Makes a Good SegAgent?}
\label{subsec:what}

Now that we have obtained the annotation trajectory dataset $D_{traj}$, in theory, we can fine-tune any MLLM to become a SegAgent. 
Here, we identify that HLMAT primarily relates to two main competencies of an MLLM:
\begin{itemize}
    \item \textbf{Coarse-Grained Localization Ability Based on Text}: This capability is primarily demonstrated when the MLLM executes the first or initial actions, where it must use the text prompt and global image context to locate the approximate position of the object. 
    This is especially critical for complex or ambiguous prompts, such as "the second black motorcycle on the right." If the model fails in this initial localization, the final IoU score will likely be very low. 
    This skill is closely related to tasks commonly known as Visual Grounding or Referring Expression Comprehension (REC), where previous work \cite{bai2023qwen,wang2024qwen2} has shown that many different architectures of MLLMs can achieve good performance when trained on sufficient data. 
    The only difference in our setup is that while other tasks typically output bounding box coordinates, our task supports point-based coordinates. 
    Our experiments further reveal that the initial action can be represented as either a box or a point
    However, this localization ability itself is not the primary focus of our work, as it has already been widely studied and optimized.     
    \item \textbf{Fine-Grained Pixel Understanding and Mask Refinement}: This capability is demonstrated when the MLLM performs subsequent actions, where it must accurately adjust the mask boundaries based on the current mask and local image information to gradually improve mask quality. 
    This process requires the model to have an exceptionally detailed perception and understanding of both the input image and mask, as well as the ability to make precise localization decisions during output. 
    Few studies have focused on this fine-grained pixel understanding of MLLMs, making this a key focus of our research.
\end{itemize}
To better investigate the second capability, a complex and high-quality segmentation dataset is essential. Here, complexity refers to the requirement that the dataset should necessitate relatively long annotation trajectories to reach the predefined threshold $\tau_{stop}$. High quality refers to minimal noise in annotations, ensuring accuracy at the trajectory endpoints.
Our analysis of the commonly used Refcoco dataset revealed that it lacks the necessary complexity and annotation quality (see \cref{subsec:datasets}). Consequently, we select two alternative datasets that better meet these criteria: DIS5K~\cite{qin2022highly} and ThinObject5K~\cite{liew2021deep}. Our further exploration will primarily focus on these datasets.

\subsection{Policy Improment}
\label{subsec:policy}
Training the MLLM with the annotation trajectory dataset $D_{traj}$ generated in \cref{subsec:trajectories}  equips the model with foundational capabilities for the HLMAT task. However, because these generated trajectories are nearly optimal and lack diversity, the model's generalization and robustness remain constrained.
It is worth noting that our framework can directly interact with the environment, allowing the MLLM to engage in online learning or offline reinforcement learning (RL) to continually refine its policy through interaction. In this paper, we choose to implement and explore a straightforward RL-style policy improvement method, specifically the StaR \cite{zelikman2022star} algorithm.

\begin{algorithm}[t]
\small
    \caption{SegAgent Policy Improvement with StaR+}
    \label{alg:star+}
    \begin{algorithmic}[1]
    \State \textbf{Input:} SegAgent model trained generated trajectory dataset $D_{\text{traj}} = \{(I^i, M_{\text{target}}^i, P^i, \mathcal{T}^i)\}_{i=1}^m$
    \State $S_0 \gets \text{SegAgent}$
    \Statex \hspace{\algorithmicindent} \textcolor{blue}{\textit{// Initialize SegAgent model}}
    \State $D_0 \gets D_{\text{traj}}$
    \Statex \hspace{\algorithmicindent} \textcolor{blue}{\textit{// Initialize trajectory dataset}}
    \For{$n = 1$ to $N$}
        \State $\hat{\mathcal{T}}^i \gets S_{n-1}(I^i, M_{\text{target}}^i, P^i)$ for all $i \in [1, m]$ 
        \Statex \hspace{\algorithmicindent} \textcolor{blue}{\textit{// Generate trajectories}}
        \State $D_n \gets \{(I^i, M_{\text{target}}^i, P^i, \textbf{Refine}(\hat{\mathcal{T}}^i)) \}_{i=1}^m$
        \Statex \hspace{\algorithmicindent} \textcolor{blue}{\textit{// Filter and correct trajectories based on reward function and $F_{\text{sim}}$}}
        \State $S_n \gets \text{train}(S_0, D_0 \cup D_n)$ 
        \Statex \hspace{\algorithmicindent} \textcolor{blue}{\textit{// Fine-tune on combined dataset}}
    \EndFor
    \end{algorithmic}
\end{algorithm}
As shown in \Cref{alg:star+}, we start with a trained SegAgent model $S_0$ and the annotation trajectory dataset $D_{traj}$. 
In each iteration, we first use $S_{n-1}$ to perform rollouts on the training images, generating a set of new  trajectories $\hat{\mathcal{T}}^i$.
 Next, we filter and refine these trajectories based on the reward function and $F_{\text{sim}}$, producing an updated trajectory dataset $D_n$. 
Finally, we merge $D_n$ with $D_{traj}$ and fine-tune the model on this combined dataset to obtain an improved SegAgent model $S_n$.
Unlike the original StaR algorithm (see comparison in Appendix \cref{alg:star}), which filters based on the overall correctness of the trajectory, our StaR+ method calculates the reward change for each individual action.
We retain actions that increase the reward, while for actions that decrease the reward, we replace them with actions generated by $F_{sim}$ as corrections.

\subsection{Process Reward Model and Tree Search}
\label{subsec:PRM}
Process-supervised Reward Models \cite{lightman2023let} (PRM) refer to the techniques where a reward model is trained to evaluate the reward at each step of a multi-step task. PRM can guide the training process by integrating with approaches like RLHF \cite{ouyang2022training} or reject sampling \cite{yuan2024rrhf}. Previous work \cite{ma2023let} has also shown that PRM can be used as guidance during the inference stage to enhance the model's search and reasoning capabilities.

In our work, since the reward at each step can be directly obtained from the environment during training, PRM is unnecessary for the training phase.  Therefore, we primarily explore the application of PRM techniques during the inference stage.
There are two main reasons motivating our use of PRM techniques:

\begin{itemize}
    \item \textbf{Providing a Stop Signal}: We use PRM to provide a stop signal for the task. An alternative approach would be to add a stop action to the action space, but this would increase the complexity of the action space, thereby increasing the difficulty of model training. This is especially challenging when $M_{target}$ contains noise, as it becomes difficult to determine the exact stopping point, making it hard for the model to generate an accurate stop action. By directly predicting the reward, PRM allows the model to decide when to stop without adding complexity to the action space.

    \item \textbf{Enhancing Robustness in Complex Scenarios}: In complex scenarios, since the model cannot backtrack or undo actions, executing an inaccurate or incorrect action at any step can adversely affect subsequent actions, leading to lower quality in the final result. With the help of PRM, we can combine the model with advanced search strategies, such as Breadth-First Search (BFS) and Monte Carlo Tree Search (MCTS), to improve the model’s robustness in complex situations.
\end{itemize}

\begin{algorithm}[t]
\small
    \caption{PRM with Heuristic Greedy Search}
    \label{alg:PRM2}
    \begin{algorithmic}[1]
    \State \textbf{Input:} SegAgent model $S$, image $I$, text prompt $P$
    \State $M_0 \gets \text{initialize\_mask}(I)$
    \Statex \hspace{\algorithmicindent} \textcolor{blue}{\textit{// Initialize the mask}}
    \State $R_0 \gets \text{PRM}(S, I, M_0, P)$
    \Statex \hspace{\algorithmicindent} \textcolor{blue}{\textit{// Predict the initial reward}}
    \For{$t = 1$ to $T$}
        \State $[a^1, a^2, ..., a^K] \gets \text{get\_candidate}(S, I, M_{t-1}, P)$
        \Statex \hspace{\algorithmicindent} \textcolor{blue}{\textit{// Get $K$ candidate actions}}
        \For {$i = 1$ to $K$}
            \State $M^i \gets F_{iter}(I, M_{t-1}, a^i)$
            \State $R^i \gets \text{PRM}(S, I, M^i, P)$
        \EndFor
        \State $\text{best\_index} \gets \text{argmax}(R^1, R^2, ..., R^K)$
        \Statex \hspace{\algorithmicindent} \textcolor{blue}{\textit{// Get the best action guided by PRM}}
        \State $M_t \gets M^{\text{best\_index}}$
        \Statex \hspace{\algorithmicindent} \textcolor{blue}{\textit{// Predict the next mask}}
        \State $R_t \gets R^{\text{best\_index}}$
        \Statex \hspace{\algorithmicindent} \textcolor{blue}{\textit{// Predict the next reward}}
    \EndFor
    \State $\text{best\_index} \gets \text{argmax}(R_0, R_1, ..., R_T)$
    \Statex \hspace{\algorithmicindent} \textcolor{blue}{\textit{// Get the best step}}
    \State \textbf{return} $M_{\text{best\_index}}$
    \end{algorithmic}
\end{algorithm}

We implement PRM by asking the MLLM perform an additional text prediction, where the MLLM learns to predict the mIoU score of the current state before generating the next action, for example, ``Current mIoU: 0''. 
Based on this, we design a simple heuristic greedy search strategy, as shown in \Cref{alg:PRM2}. At each step, we first use Multinomial Sampling to generate $K$ candidate actions. For each action, we perform one interaction with the segmentation model to obtain a new mask and calculate the reward using PRM. Finally, we select the action with the highest reward as the next action. This process continues until the maximum number of steps $T$ is reached or until the reward converges. We then return the mask with the highest reward as the final output.
A more detailed illustration can be found in the appendix. Note that, in the algorithm description, we omit the historical actions included in the input of $F_{iter}$ for simplicity.
It is worth noting that more complex search strategies, such as BFS and MCTS, are also feasible. However, there exists a trade-off between computational complexity and performance. Our experiments demonstrate that the simplest and most efficient heuristic greedy search strategy already provides significant performance improvements. We leave the exploration of more complex search strategies for future work.

\begin{table*}[ht]
    \centering
    \caption{Comparison of methods on RES dataset. %
    We indicate which models use SAM. ``SClick'' denotes the use of SimpleClick as the interactive segmentation model.}
    \label{tab:main_result}
        \vspace*{-0.3cm}
    {\resizebox{0.81\textwidth}{!}{
    \begin{tabular}{l c ccc ccc cc}
    \toprule
    \multirow{2}{*}{\textbf{Method}} & \multirow{2}{*}{\textbf{Venue}} & \multicolumn{3}{c}{\textbf{refCOCO}} & \multicolumn{3}{c}{\textbf{refCOCO+}} & \multicolumn{2}{c}{\textbf{refCOCOg}} \\ 
    \cmidrule(lr){3-5}\cmidrule(lr){6-8}\cmidrule(lr){9-10}
    && val & testA & testB & val & testA & testB & val(U) & test(U) \\
    \midrule
    \emph{traditional methods} \\
    MAttNet~\cite{yu2018mattnet} & CVPR'18 & 56.51 & 62.37 & 51.70 & 46.67 & 52.39 & 40.08 & 47.64 & 48.61 \\
    LAVT~\cite{yang2022lavt} & CVPR'22 & 72.7 & 75.8 & 68.8 & 62.1 & 68.4 & 55.1 & 61.2 & 62.1 \\
    CRIS~\cite{wang2022cris} & CVPR'22 & 70.5 & 73.2 & 66.1 & 65.3 & 68.1 & 53.7 & 59.9 & 60.4 \\
    PolyFormer-L~\cite{liu2023polyformer} & CVPR'23 & 76.94 & 78.49 & 74.83 & 72.15 & 75.71 & 66.73 & 71.15 & 71.17 \\
    X-Decoder~\cite{zou2023generalized} & CVPR'23 & - & - & - & - & - & - & 64.6 & - \\
    SEEM~\cite{zou2024segment} & NeurIPS'23 & - & - & - & - & - & - & 65.7 & - \\
    \midrule
    \emph{LLM based methods} \\
    \color{black}LISA(SAM)~\cite{lisa} & CVPR'24 & 74.9 & 79.1 & 72.3 & 65.1 & 70.8 & 58.1 & 67.9 & 70.6 \\
    \color{black}PixelLM~\cite{ren2024pixellm} & CVPR'24& 73.0 & 76.5 & 68.2 & 66.3 & 71.7 & 58.3 & 69.3 & 70.5 \\
    \color{black}PerceptionGPT~\cite{pi2023perceptiongpt} & CVPR'24& 75.1 & 78.6 & 71.7 & 68.5 & 73.9 & 61.3 & 70.3 & 71.7 \\
    \color{black}GSVA(SAM)~\cite{xia2023gsva} & CVPR'24& {77.2} & 78.9 & {73.5} & 65.9 & 69.6 & 59.8 & 72.7 & 73.3 \\
    SAM4MLLM(Qwen)~\cite{chen2025sam4mllm} & ECCV'24 & 77.1 & {80.9} & 72.5 & {71.5} & {76.8} & {64.7} &  {74.5} & {75.2} \\
    Qwen box~\cite{bai2023qwen} + SAM & - & 71.79 & 75.20 & 67.29 & 65.39 & 71.62 & 59.12 & 66.93 & 68.94 \\
    \midrule
    \emph{Our methods} \\
    SegAgent-LLaVA+SAM & \multirow{4}{*}{{this work}} & 79.20 & 81.44 & 75.72 & 71.53 & 76.68 & 65.44 & 74.80 & 74.90 \\
    SegAgent-Qwen+SAM  & & 78.01 & 80.34 & 74.98 & 70.86 & 75.52 & 65.75 & 74.49 & 74.62 \\
    SegAgent-LLaVA+SClick && 77.81 & 80.03 & 74.12 & 66.73 & 71.16 & 59.89 & 70.45 & 71.25 \\
    SegAgent-Qwen+SClick & & 79.69 & 81.35 & 76.57 & 72.49 & 75.80 & 66.89 & 75.11 & 75.20 \\
    \bottomrule
    \end{tabular}
    }}
        \vspace*{-0.6cm}
\end{table*}

\section{Experiments}
\label{sec:experiments}

\subsection{Evaluation Protocol}
\label{subsec:protocol}

As mentioned earlier, HLMAT serves as a novel evaluation protocol for assessing and advancing MLLMs' fine-grained pixel understanding capabilities. It also provides a framework to explore MLLMs as vision-centered, multi-step decision-making agents. We select two representative MLLMs as initialization models for SegAgent: a linear projector-based model, LLaVA-v1.5-7B \cite{liu2024llava,liu2024improved}, and a stronger Q-former-based model, Qwen-VL-7B. We fine-tune each MLLM on the generated trajectory dataset for 2 epochs, keeping the image encoder frozen while allowing the LLM and projector layers to update. We use DeepSpeed’s Zero2 mode for multi-GPU training, with all training experiments conducted on 8 NVIDIA 80GB GPUs.

Other training parameters follow the default settings provided by the official repositories; details can be found in the Appendix. We used two main interactive segmentation models for this task: SimpleClick \cite{liu2023simpleclick} and SAM \cite{sam}. SimpleClick supports point input only, while SAM also supports box input. During trajectory generation, we created trajectories for both segmentation models. As discussed in \cref{subsec:what}, SegAgent’s first action can be either a point or a box. Our preliminary experiments demonstrated that SAM achieves better performance when provided with an additional box input. Therefore, unless otherwise specified, we include box input when using SAM in subsequent experiments.

We conduct experiments in two settings: the widely adopted Referring Expression Segmentation (RES) and our newly proposed High-quality Referring Expression Segmentation (HRES). 

For inference evaluation, we set a fixed maximum number of steps $T$ and, following LISA \citep{lisa}, used cumulative IoU (cIoU) as the evaluation metric. For RES, due to its relatively simple masks, we set $T=7$ and $K=1$ in \cref{alg:PRM2}. For HRES, with more complex segmentation requirements, we set $T=11$ and $K=3$.

\subsection{Datasets}
\label{subsec:datasets}

We first conduct a fair comparison between our method and other state-of-the-art (SOTA) methods on the Referring Expression Segmentation (RES) dataset to demonstrate the effectiveness of our approach. Subsequently, we extend our method to more complex scenarios by experimenting on the proposed High-quality Referring Expression Segmentation (HRES) dataset. We begin by introducing the basic information of these two datasets and analyzing their characteristics to highlight the necessity of using the HRES dataset.

\textbf{Referring Expression Segmentation (RES)}: The RES dataset is composed of three subsets: refCOCO~\cite{yu2016modeling}, refCOCO+~\cite{yu2016modeling}, and refCOCOg~\cite{mao2016generation}. All three datasets are based on images from the COCO dataset, with differences in the types of referring expressions provided. RefCOCO is the earliest dataset, featuring relatively simple descriptions. Compared to RefCOCO, RefCOCO+ prohibits annotators from using location-based descriptions, making the descriptions more focused on object appearance and attributes. RefCOCOg includes longer descriptions with more detailed information. However, the mask quality in refCOCO(+/g) is relatively low, with many masks containing significant noise. This makes it challenging to generate long and precise annotation trajectories. On the other hand, the complexity of the datasets does not meet the requirements for the multi-step iteration in HLMAT. In most cases, only 1-2 steps are sufficient to complete the task. 

To address these limitations, we introduce a new dataset, the High-quality Referring Expression Segmentation (HRES) dataset, designed to facilitate further exploration and better assess the performance of MLLMs under more complex conditions.

\begin{figure}
    \centering
    \includegraphics[width=0.7\linewidth]{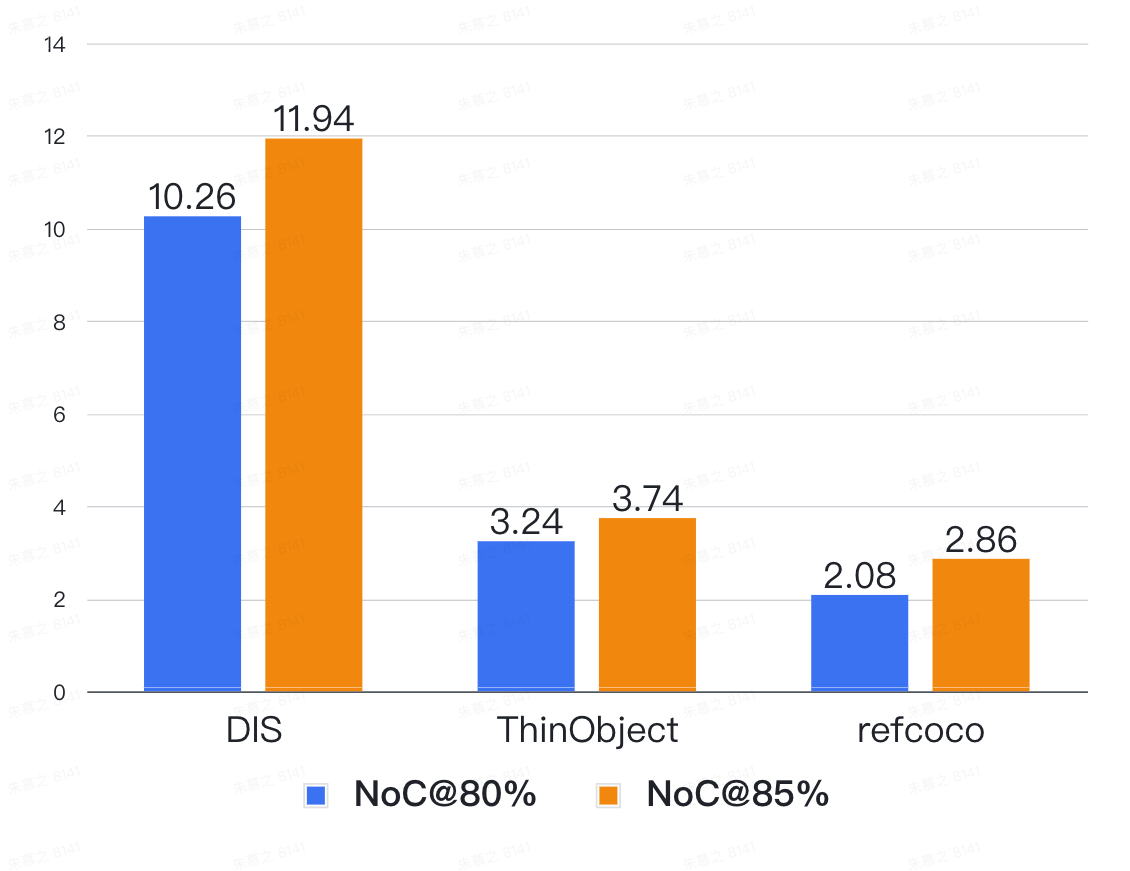}
    \vspace*{-0.5cm}
    \caption{%
    Comparison of dataset complexity.}
    \label{fig:dataset}
    \vspace*{-0.7cm}
\end{figure}
\textbf{High-quality Referring Expression Segmentation (HRES)}: Our proposed HRES dataset features higher annotation quality, with more detailed and complex masks, which require additional steps to complete. The data in HRES is derived from the HQSeg-44K dataset proposed by HQ-SAM~\cite{ke2024segment}. We selected two subsets, DIS5K~\cite{qin2022highly} and ThinObject5K~\cite{liew2021deep}, as they contain texts or category descriptions with minimal ambiguity.The DIS5K (Dichotomous Image Segmentation) subset consists of high-resolution images and high-quality binary segmentation masks, specifically designed to handle challenging segmentation tasks. It includes 5,470 images spanning 22 groups and 225 fine-grained object categories. The ThinObject5K subset focuses on objects with thin, elongated structures, such as insect legs and bicycle spokes, which are typically challenging for segmentation models. 

As shown in \cref{fig:dataset}, we use the number of clicks (required to reach the target IoU) as a measure of dataset complexity. The HRES dataset is significantly more complex than the RES dataset, requiring more steps to achieve the target IoU. 
Furthermore, we provide a qualitative analysis of the annotation quality in the appendix to further illustrate the necessity of the HRES dataset.

\subsection{Main Results}
\label{subsec:mains}
In this section, we present the main results of SegAgent on the HLMAT task, comparing it with other state-of-the-art (SOTA) methods on RES datasets in \cref{tab:main_result}. We compare SegAgent with both traditional and LLM-based methods. Our results demonstrate that SegAgent achieves competitive IoU scores, particularly SegAgent-LLaVA+SAM and  SegAgent-Qwen+SClick, which outperforms all other methods except for GLaMM.

We also implement a simple baseline method, Qwen box + SAM, by leveraging Qwen-VL’s grounding capability to obtain a bounding box, which is then passed to SAM to generate the mask. For fair comparison, our SegAgent-LLaVA(Qwen)+SAM also uses the same box as the initial action to evaluate SegAgent’s ability to refine the mask. Results show that SegAgent-LLaVA(Qwen)+SAM outperforms the Qwen box + SAM baseline across all datasets, indicating that our approach effectively refines the mask.

An interesting observation is that although Qwen is trained on a larger dataset and generally outperforms LLaVA on standard vision-language tasks, LLaVA surpasses Qwen in mask refinement capability. We attribute this to Qwen's reliance on the Q-former structure, which appears to struggle with pixel-level tasks. Recent studies \cite{cha2024honeybee,yao2024deco}  have reported similar limitations, suggesting that the vanilla QFormer might lose spatial locality during semantic abstraction. We envision HLMAT as a future protocol for assessing MLLM’s fine-grained pixel-level visual abilities, inspiring further exploration into improving the pixel-level understanding abilities of these models.

Additionally, when using SimpleClick as the interactive segmentation model, we observe that SegAgent-Qwen significantly outperforms SegAgent-LLaVA. This aligns with the first capability discussed in \cref{subsec:what}, as it plays a more dominant role here. Qwen’s pretraining on large-scale visual grounding tasks gives it an advantage in locating the target during the initial interactions, leading to better overall performance.

\begin{figure}
    \centering
    \includegraphics[width=0.7\linewidth]{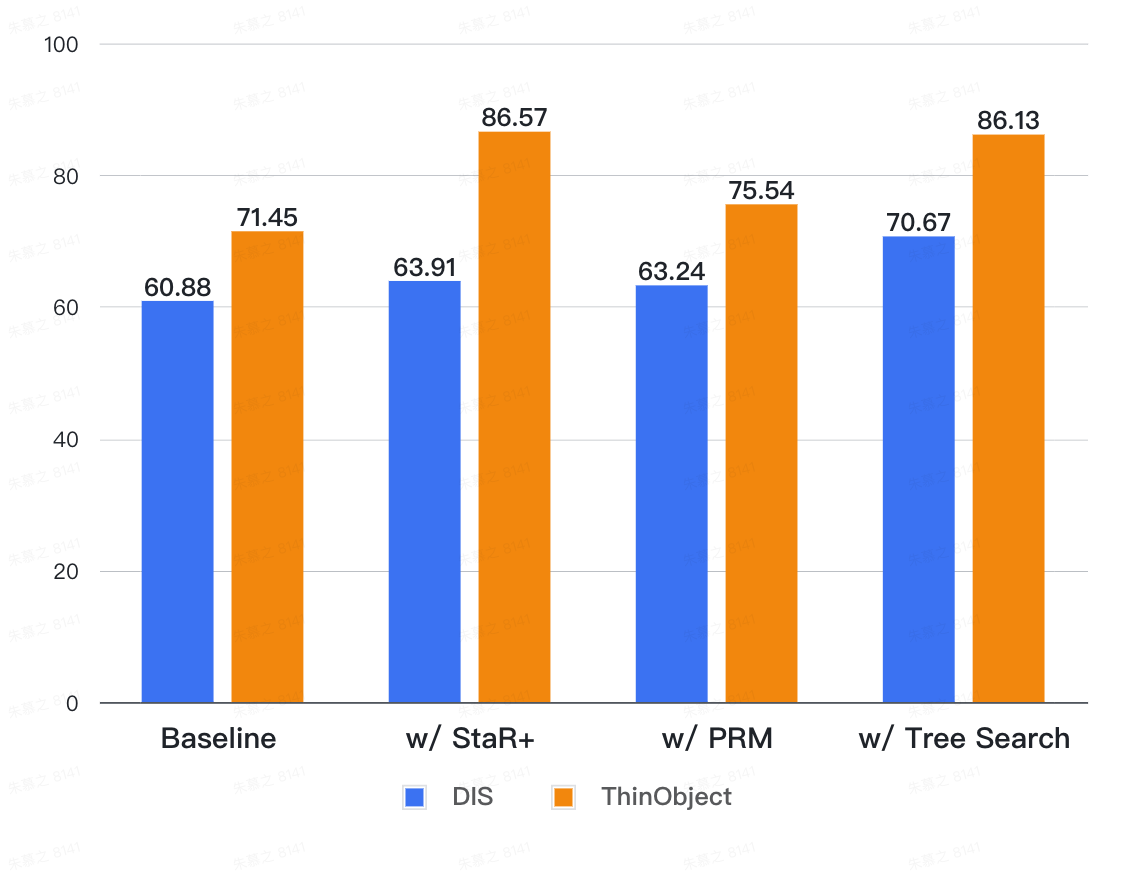}
    \vspace*{-0.5cm}
    \caption{Comparison of different strategies on the HRES dataset.}
    \label{fig:enter-label}
    \vspace*{-0.7cm}
\end{figure}

\subsection{Exploration on More Complex Scenarios}
\label{subsec:complex}
As discussed in \cref{subsec:datasets}, we validate our proposed Policy-Improvement strategy, as well as the effectiveness of PRM and Tree Search, on a more complex and accurate dataset. Here, we take the model fine-tuned on generated trajectories with SFT (Supervised Fine-Tuning) and apply a fixed-step greedy decoding during testing as our baseline.

\textbf{Policy-Improvement Strategies}: As mentioned in \cref{alg:star+}, we use the baseline model trained on generated trajectories, and then conduct policy improvement using StaR+ to enhance the model's performance. 
Although we only use a single round of StaR+, we find that this strategy significantly improves the model's performance. The improvement is particularly pronounced on the ThinObject5K dataset, where our model achieves an IoU of 86.57 compared to the baseline IoU of 71.45.

\textbf{PRM and Tree Search}: 
As shown in the figure, introducing PRM as a stopping signal improves model performance compared to the baseline, which relies on fixed-step greedy decoding. Specifically, with PRM alone (equivalent to setting \( K=1 \)), the model performance on DIS increases by 2.36 and on ThinObject by 4.09. Furthermore, when we set \( K=3 \) and apply our proposed Tree Search algorithm, the model performance is further enhanced, achieving an additional increase of 7.43 on DIS and 10.59 on ThinObject compared to using only PRM. These results align with our hypothesis that Tree Search effectively prevents the model from getting trapped in local optima when making uncertain actions, thus boosting performance in complex scenarios.

\section{Conclusion}
\label{sec:conclusion}

In this work, we introduced the Human-Like Mask Annotation Task (HLMAT), a novel segmentation paradigm that models MLLMs as multi-step decision-making agents, imitating human annotation paths in interactive segmentation. Through HLMAT, we assess and enhance MLLMs' pixel-level understanding without requiring additional model structures. We developed SegAgent by fine-tuning MLLMs on human-like trajectories and demonstrated its competitive performance across various segmentation tasks, including mask refinement and annotation filtering. Additionally, we introduced the high-quality HRES dataset and showed that decision-enhancement techniques like StaR+ and tree search with PRM can further boost performance. Our study paves the way for future exploration of vision-centered, multi-step decision-making with MLLMs.
\section*{Acknowledgement}
This work is partially supported by the National Key R\&D Program of China(NO.2022ZD0160101) and the National Natural Science Foundation of China (No.\ 62206244). This work is also supported by Ant Group Research Intern Program. 
{
    \small
    \bibliographystyle{ieeenat_fullname}
    \bibliography{main}
}

\clearpage
\maketitlesupplementary

\appendix

\section{Discussion}

\subsection{Computational complexity Trade-off}
We agree that error accumulation is a potential issue, and longer sequences may lead to worse results if we perform fixed-length greedy decoding. However, the proposed PRM can effectively alleviate this issue. The experimental results are as follows. 

\begin{table}[h]
  \vspace{-4mm}
  \centering
  \footnotesize %
  \setlength{\tabcolsep}{3pt} %
  \renewcommand{\arraystretch}{0.9} %
  \captionsetup{aboveskip=2pt, belowskip=2pt} %
  \caption{Performance comparison under different steps.}
  \label{tab:complexity_trade_off}
  \begin{tabular}{lcccc}
  \toprule
  \# Steps & 1 & 3 & 5 & 7 \\ \midrule
  w/o PRM & 71.53 & 73.67 & 73.88 & 68.22 \\
  w/ PRM  & 71.53 & 72.98 & 75.21 & 75.43 \\ \bottomrule
  \end{tabular}
  \vspace{-4mm}
\end{table}

\subsection{Limitations}

\textbf{Failure cases.} Failures mainly arise from (1) incorrect target localization in the first step and (2) inaccurate coordinate outputs in refinement steps, particularly at boundaries (wrong point reaching the background area), resulting in the wrong mask.(see \cref{fig:predict})

We believe that the current bottleneck limiting the model's performance lies mainly in the fine-grained pixel-level localization ability of MLLMs.
On the HRES dataset, the main reason for the model's failure is the inaccurate location of the output points, such as the wrong point being outside the object's boundary.

\section{More Details about algorithm}
\subsection{Trace Generation}
In Section 4.1 of the main text, we introduced how $F_{\text{sim}}$ is used to simulate user trajectories. Here, we provide a more detailed illustration of this process in the form of Python pseudocode (see \Cref{alg:code}).
\begin{algorithm}[h]
    \caption{Pseudo code of trace generation}
    \label{alg:code}
    \definecolor{codeblue}{rgb}{0.25,0.5,0.5}
    \lstset{
        backgroundcolor=\color{white},
        basicstyle=\fontsize{7.2pt}{7.2pt}\ttfamily\selectfont,
        columns=fullflexible,
        breaklines=true,
        captionpos=b,
        commentstyle=\fontsize{7.2pt}{7.2pt}\color{codeblue},
        keywordstyle=\fontsize{7.2pt}{7.2pt},
    }
    \begin{lstlisting}[language=python]
        # n: Maximum number of clicks 
        # image: The image to be segmented
        # gt_mask: Ground truth mask of current image
        # pred_mask: The predicted mask according to clicks in click_list
    
        click_list = []
        pred_mask = np.zeros_like(image)
    
        # Iterate n times
        for i in [1, n]:
            fn = gt_mask & (~pred_mask) # false negative
            fp = (~gt_mask) & pred_mask # false positive
            fn_dist = cv2.distanceTransform(fn_mask)
            fp_dist = cv2.distanceTransform(fp_mask)
            fn_max_dist = max(fn_dist)
            fp_max_dist = max(fp_dist)
    
            if (fn_max_dist > fp_max_dist): # Next click should be positive
                click_y, click_x = np.where(fn_dist == fn_max_dist)
                is_positive = 1
            else: # Next click should be negative
                click_y, click_x = np.where(fp_dist == fp_max_dist)
                is_positive = 0
    
            click_list.append((click_x, click_y, is_positive))
            pred_mask = model.predict(image, click_list)
    
\end{lstlisting}
\end{algorithm}

To further enhance the diversity of annotation trajectories and improve SegAgent's mask refinement capabilities, we introduced additional initial states beyond the empty mask. These include masks generated from bounding boxes and masks created by random sampling points.

\subsection{Comparision with StaR}
As mentioned in Section 4.3 of the main text, we have introduced the differences between our method StaR+ and the original StaR algorithm \cite{zelikman2022star}. 
Here we first present the original StaR algorithm (see \Cref{alg:star}). 
In addition to the differences mentioned in the main text, another difference is that the original StaR algorithm will only fine-tune the model with the newly generated trajectory dataset $D_n$, while our StaR+ will fine-tune the model with the merged dataset of $D_{\text{traj}}$ and $D_n$. This is because the original StaR algorithm is designed for reasoning tasks that lack a simulate function $F_{\text{sim}}$ to generate trajectories. In our task, $D_{\text{traj}}$ retains a lot of useful information because it generates approximately optimal trajectories. Therefore, we merge it with $D_n$ to fine-tune the model.
\begin{algorithm}[H]
    \caption{SegAgent Policy Improvement with StaR}
    \label{alg:star}
    \begin{algorithmic}[1]
    \State \textbf{Input:} The SegAgent model trained on the generated trajectory dataset $D_{\text{traj}} = \{(I^i, M_{\text{target}}^i, P^i, \mathcal{T}^i)\}_{i=1}^m$
    \State $S_0 \gets \text{SegAgent}$
    \Statex \hspace{\algorithmicindent} \textcolor{blue}{\textit{// Initialize the SegAgent model}}
    \State $D_0 \gets D_{\text{traj}}$
    \Statex \hspace{\algorithmicindent} \textcolor{blue}{\textit{// Initialize the trajectory dataset}}
    \For{$n = 1$ to $N$}
        \State $\hat{\mathcal{T}}^i \gets S_{n-1}(I^i, M_{\text{target}}^i, P^i)$ for all $i \in [1, m]$ 
        \Statex \hspace{\algorithmicindent} \textcolor{blue}{\textit{// Perform trajectory generation}}
        \State $D_n \gets \{(I^i, M_{\text{target}}^i, P^i, \textbf{Filter}(\hat{\mathcal{T}}^i)) \mid i \in [1, m] \}$ 
        \Statex \hspace{\algorithmicindent} \textcolor{blue}{\textit{// Filter trajectories based on the reward function}}
        \State $S_n \gets \text{train}(S_0, D_n)$ 
        \Statex \hspace{\algorithmicindent} \textcolor{blue}{\textit{// Fine-tune the model on the filtered dataset}}
    \EndFor
    \end{algorithmic}
\end{algorithm}

\subsection{More Details about Process Reward Model and Tree Search}
\begin{figure*}[t]
    \centering
    \includegraphics[width=1\linewidth]{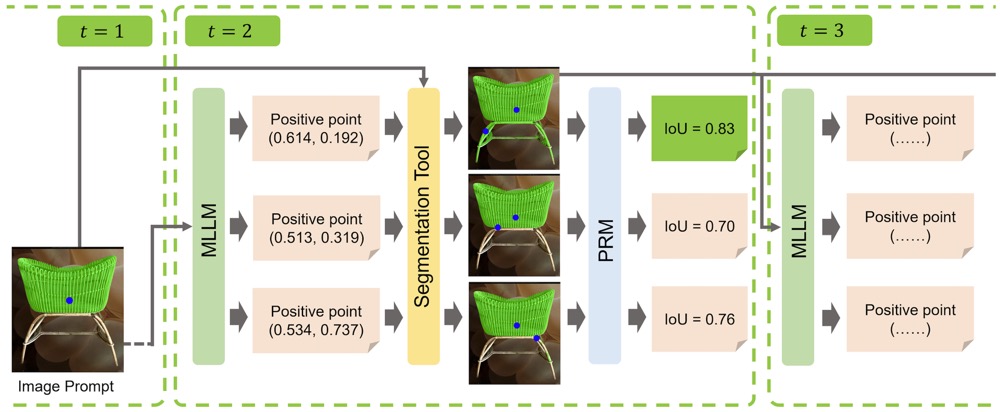}
    \vspace*{0.1cm}
    \caption{%
    An illustrative example of PRM-guided tree search. The model predicts the reward at each step and selects the action with the highest reward to generate the next mask.}
    \vspace*{0.3cm}
    \label{fig:prm}
\end{figure*}

In Section 4.4 of the main text, we have shown the process of PRM-guided tree search in the form of pseudocode. To facilitate readers' understanding, we further illustrate this process in \cref{fig:prm}

\begin{figure*}[ht]
    \centering
    \setlength{\fboxrule}{0.9pt}
    \fbox{
        \parbox{0.9\textwidth}{
            \textbf{Prompt Design for SegAgent}\\
            \footnotesize
            You are a highly skilled segmentation annotator. We have provided you with an image and an initial mask marked by a \textbf{semi-transparent green mask} that roughly covers the object described below. Your task is to refine this mask to make it as accurate as possible. Based on the given image and the mask, perform the following actions: \\
            \\
            \textbf{1. Positive Point (x, y):} \\
            Add a positive point if any part of the object is not covered by the mask. This will expand the mask to include the missing area.  
            \textit{Example:} Add a positive point on any corner or edge of the object that the mask does not cover. \\
            \\
            \textbf{2. Negative Point (x, y):} \\
            Add a negative point if an area outside the object is incorrectly included in the mask. This will refine the mask by excluding unnecessary regions.  
            \textit{Example:} Add a negative point where the mask extends into the background or any non-object area. \\
            \\

            The description of the object is as follows: \red{\textless description\textgreater}.
            }
        }
    \caption{The prompt provides detailed instructions for refining a segmentation mask with three possible actions: adding a positive point, adding a negative point. The \red{red} part indicates user-specific input, such as object descriptions. }
    \label{fig:prompt-design}
\end{figure*}
\section{Implementation Details}

\subsection{Model Architecture and Hyperparameters}
For SegAgent-LLaVA, we initialized the model with project weights provided by \cite{yuan2024osprey}. Subsequently, we performed a second-stage fine-tuning using the annotation trajectories generated in our framework. Following the implementation in \cite{yuan2024osprey}, we adopted a ConvNeXt-L \cite{liu2022convnet} CLIP model as the vision encoder, extracting image features from the "res4" stage.
The model was trained using the AdamW \cite{adamw} optimizer with a learning rate of $1 \times 10^{-5}$ and a cosine annealing scheduler \cite{loshchilov2016sgdr} for two epochs. We set the batch size to 16. During both training and inference, input images were resized to $512 \times 512$. The maximum sequence length was set to 2048 tokens.

For SegAgent-QWen, we initialized the model using the Qwen-VL-Chat weights provided in the official implementation \cite{bai2023qwen}. Fine-tuning was conducted using the full-parameter fine-tuning script provided by the authors, with only the ViT module frozen. 
Specifically, input images were resized to $448 \times 448$, and 256 queries were used for the vision-language adapter. The model was trained using the AdamW optimizer with a learning rate of $1 \times 10^{-5}$, a cosine decay learning rate schedule, and a batch size of 128 for two epochs. 
The maximum sequence length was set to 2048 tokens.

For SAM \cite{sam} and SimpleClick \cite{liu2023simpleclick}, we used the official pre-trained weights provided by their respective repositories. Both models are based on a ViT-large architecture.

\subsection{Prompt Design}
Here we provide a detailed introduction to the specific design of the input prompt $P$ for MLLMs, as shown in \cref{fig:prompt-design}. The design of the prompt is to guide the model to generate more accurate annotations, including two operations: adding a positive point, adding a negative point. Adding a positive point is to expand the mask, and adding a negative point is to shrink the mask.

\begin{figure*}[t]
    \centering
    \includegraphics[width=1\linewidth]{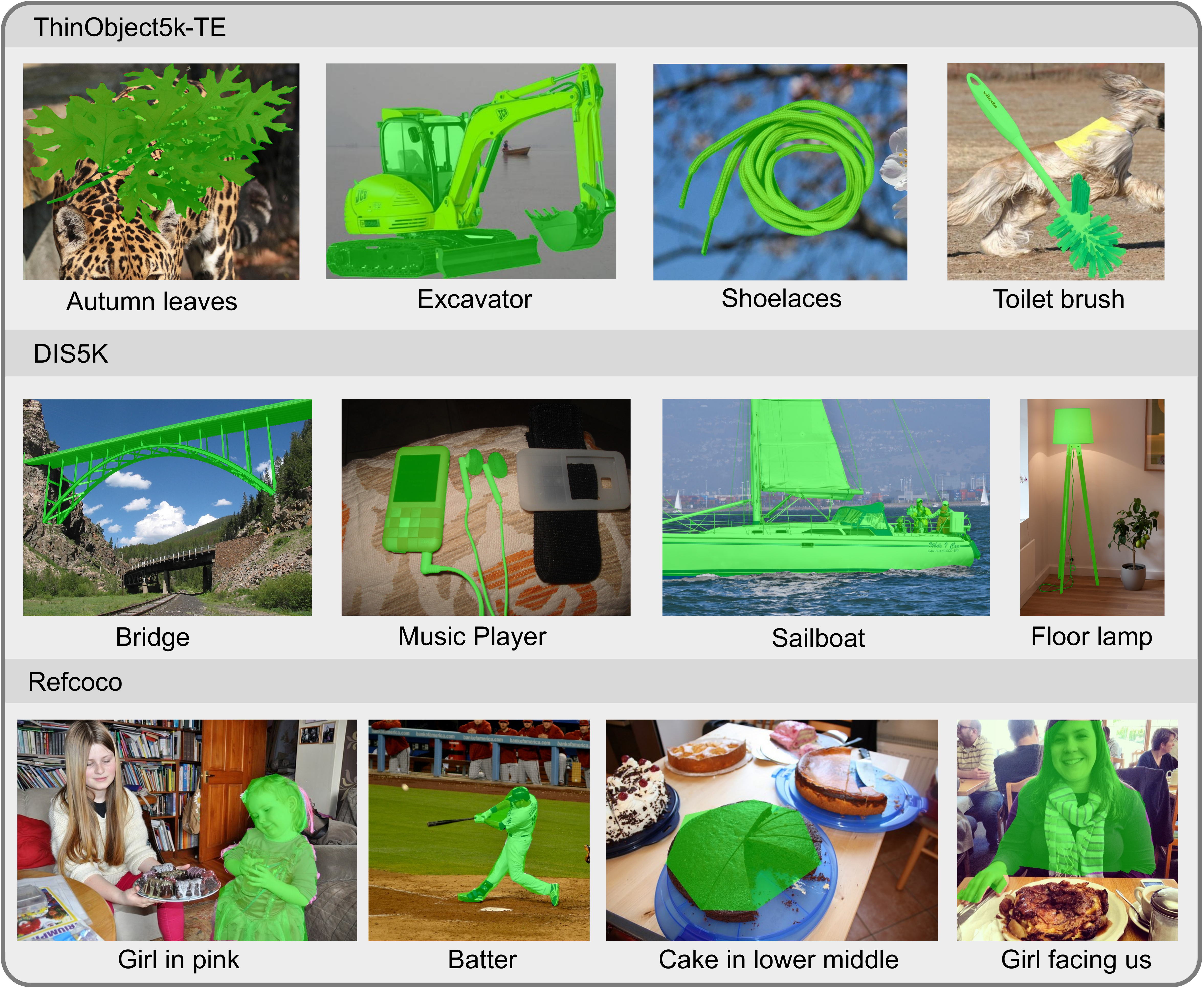}   
    \caption{\small {\bf Examples of Images and Annotations from Various Datasets.} The figure showcases representative samples from three datasets: ThinObject5k-TE, DIS5K, and RefCOCO. Each row represents a dataset, with images and corresponding annotations highlighting different objects and scenes. The annotations (green overlays) demonstrate the varying levels of detail and complexity across datasets.}

    \label{fig:dataset-visualization}
\end{figure*}

\section{Visualization Analysis}
\subsection{Comparison of Dataset Quality}
In Section 5.2 of the main text, we quantitatively analyzed the complexity of different datasets. Here we now provide a qualitative comparison of dataset quality through visualization.
\Cref{fig:dataset-visualization} illustrates examples of images and annotations from various datasets, allowing readers to gain a deeper understanding of the characteristics of each dataset.

From the visualization, we can observe that the annotation masks in ThinObject5k-TE and DIS5K are indeed more complex and precise. For instance, in the "Bridge" and "Sailboat" examples from DIS5K, the annotations exhibit intricate details such as hollow structures and fine lines. These characteristics highlight the high annotation quality and attention to detail in these datasets. 

In contrast, RefCOCO primarily focuses on scenes with people and common objects. Although the captions are longer, the annotations contain more noise. For example, while the masks roughly cover the objects, there are significant issues with mislabeling and omissions at the edges. Additionally, RefCOCO struggles to handle intricate details such as hollow regions effectively.

In summary, ThinObject5k-TE and DIS5K offer higher-quality and more complex annotations, making them better suited for evaluating and exploring SegAgent's ability to refine masks over multiple steps.
\subsection{Visualization of Predicted Trajectories}
\begin{figure*}[t]
    \centering
    \includegraphics[width=1\linewidth]{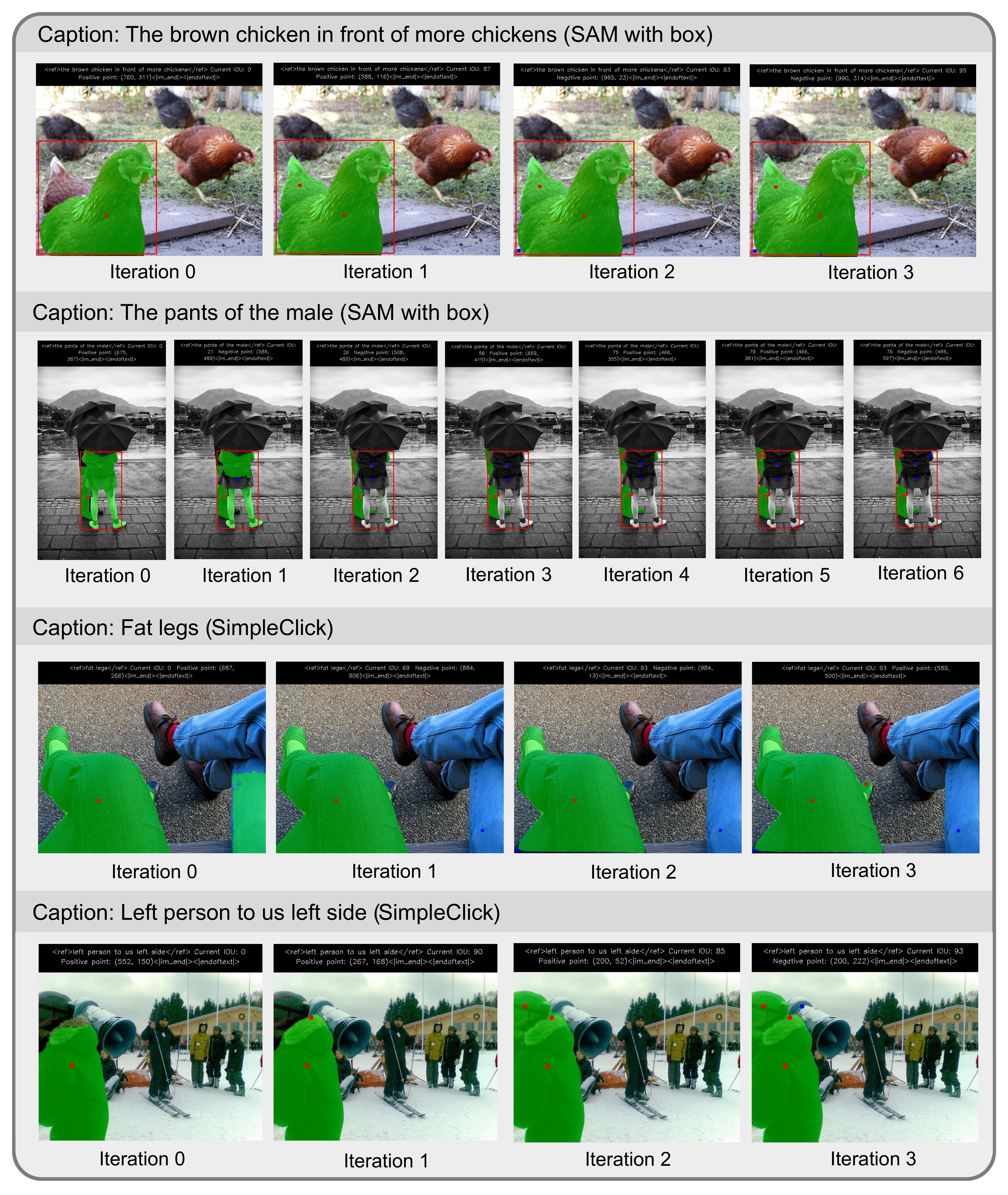}   
    \caption{\small {\bf Predicted trajectories of SegAgent using SAM and SimpleClick.} We visualize current action $a_t$ and the resulting mask 
    $M_{t+1}$ in one image. Red points represent positive points, and blue points represent negative points. }

    \label{fig:predict}
\end{figure*}

We visualized the original predicted trajectories of SegAgent, as shown in \cref{fig:predict}. Note that PRM and Tree Search were not used in this visualization.
The first two rows show the results of using Qwen-box as the first action combined with SAM for mask refinement. Although we visualized the clicks at Iteration 0, the first click was not actually input to SAM. In subsequent iterations, we used the clicks predicted by SegAgent and Qwen-box together as input to SAM.
The last two rows show the results of using only clicks as actions combined with SimpleClick for mask annotation.
It can be observed that SegAgent has indeed learned the rules of annotation and acquired an understanding of objects. It can refine masks step by step through positive and negative points.
\section{More Experiments}

\subsection{Annotation Filtering}
In Section 5.3 of the main text, we analyzed SegAgent's capabilities in mask annotation and mask refinement. Here, we further explore and demonstrate its ability in annotation filtering.

We model annotation filtering as a regression task, where the model predicts the Intersection over Union (IoU) between the current mask and the ground truth (GT) mask. This functionality is a key feature of SegAgent's PRM. In practice, by setting a reasonable threshold, we can effectively filter out low-quality masks.

To evaluate this ability, we constructed a test set based on the validation set of RefCOCO. Specifically, we generated masks of varying quality by randomly sampling positive and negative points within the GT bounding boxes. The PRM was then used to predict the  IoU of these masks.

We assessed the annotation filtering capability of the PRM using several standard regression metrics, including Mean Squared Error (MSE), Mean Absolute Error (MAE), Pearson Correlation Coefficient, and Spearman Correlation Coefficient.

\begin{table}[ht]
    \centering
    \caption{\small {\bf Evaluation of SegAgent's Annotation Filtering Ability.} Lower MAE and MSE indicate better accuracy, while higher Pearson and Spearman correlation coefficients reflect stronger agreement with ground truth IoU.}
    \label{tab:annotation-filtering}
    \resizebox{\columnwidth}{!}{
        \begin{tabular}{lcccc}
            \toprule
            \textbf{Model} & \textbf{MAE} $\downarrow$ & \textbf{MSE} $\downarrow$ & \textbf{Pearson} $\uparrow$ & \textbf{Spearman} $\uparrow$ \\
            \midrule
            SegAgent-Qwen & 6.88 & 193.98 & 0.90 & 0.87 \\
            SegAgent-LLaVA & 5.58 & 175.35 & 0.91 & 0.90 \\
            \bottomrule
        \end{tabular}
    }
\end{table}

Based on the results in \cref{tab:annotation-filtering}, SegAgent-LLaVA outperforms SegAgent-Qwen across all evaluated metrics, indicating its superior ability in annotation filtering.
These results are consistent with the analysis in Section 5.3 of the main text regarding the mask refinement capability. We hypothesize that the differences in performance may stem from the distinct model architectures. Specifically, the Q-former structure in SegAgent-Qwen might lead to some loss of detail, which could explain its slightly inferior performance compared to SegAgent-LLaVA.

However, from an overall perspective, both SegAgent-LLaVA and SegAgent-Qwen exhibit high correlation coefficients and relatively low MAE and MSE. This indicates that the PRM is highly effective in predicting the mIoU of masks, enabling the filtering of low-quality masks with strong reliability.

\subsection{Mask color}
By default, we visualize the current segmentation results using a semi-transparent green mask. Here, we further investigate the impact of mask color on segmentation performance.
\begin{table}[ht]
    \centering
    \caption{\small {\bf Evaluation of Mask Color on Segmentation Performance.}}
    \label{tab:mask-color}
    \begin{tabular}{lccc}
        \toprule
        \textbf{Mask Color} & \textbf{Green} & \textbf{Blue} & \textbf{Red} \\
        \midrule
        \textbf{mIoU} & 0.749 & 0.750 & 0.749 \\
        \bottomrule
    \end{tabular}
\end{table}
The results in Table \ref{tab:mask-color} show the impact of mask color on segmentation performance, measured by mean Intersection over Union (mIoU). 
The three tested mask colors—green, blue, and red—yield nearly identical performance.
This suggests that the choice of mask color has minimal, if any, effect on segmentation performance. The consistent mIoU across different colors indicates that the model's segmentation capability is robust to visual variations in mask color.
\subsection{Init Action}
\begin{table}[ht]
    \centering
    \caption{\small {\bf Evaluation of Different Initial Actions on SegAgent Performance.}}
    \label{tab:init-action}
    \begin{tabular}{lccc}
        \toprule
        \textbf{Initial Action} & \textbf{NO box} & \textbf{Qwen Box} & \textbf{Self Box} \\
        \midrule
        \textbf{refcoco(val)} & 77.81 & 78.01 & 77.85 \\
        \textbf{refcoco+(val)} & 70.88 & 70.86 & 70.50 \\
        \textbf{refcocog(test)} & 73.13 & 74.62 & 74.33 \\
        \bottomrule
    \end{tabular}
\end{table}
Since SAM can accept both boxes and clicks as input, we investigated the impact of different initial actions on segmentation performance.
In Table \ref{tab:init-action}, "NO box" indicates using only clicks as actions, "Qwen Box" represents using the box predicted by Qwen-VL-chat as the action, and "Self Box" denotes using the box predicted by SegAgent-Qwen itself as the action (an additional task during training).
The results indicate that the choice of initial action has a minimal impact on segmentation performance, suggesting that the model is robust to the selection of initial actions. Overall, using Qwen Box as the initial action yields slightly better performance than the other two initial actions. 
To ensure a fair comparison, we selected Qwen Box as the initial action for SAM.
\subsection{Coordinate Format}
We also investigated the representation of coordinates. For SegAgent-Qwen, we used the [0, 1000) format to represent the coordinates of bounding boxes, as Qwen itself has grounding capabilities.
For SegAgent-LLaVA, we explored whether to use integers in the range [0, 1000) to represent relative positions or decimals in the range [0, 1).
\begin{table}[ht]
    \centering
    \caption{\small {\bf Evaluation of Coordinate Format on Segmentation Performance.}}
    \label{tab:coordinate-format}
    \begin{tabular}{lcc}
        \toprule
        \textbf{Coordinate Format} & \textbf{[0, 1)} & \textbf{[0, 1000)} \\
        \midrule
        \textbf{mIoU} & 0.749 & 0.747 \\
        \bottomrule
    \end{tabular}
\end{table}
Table \ref{tab:coordinate-format} shows the impact of relative position representation on segmentation performance. The results indicate that the two coordinate formats yield nearly identical performance, suggesting that the model is robust to the choice of coordinate format. For SegAgent-LLaVA, we selected decimals in the range [0, 1) to represent relative positions.

\end{document}

%% file: preamble.tex
\newcommand{\red}[1]{{\color{red}#1}}

%% file: main.bbl
\begin{thebibliography}{82}
\providecommand{\natexlab}[1]{#1}
\providecommand{\url}[1]{\texttt{#1}}
\expandafter\ifx\csname urlstyle\endcsname\relax
  \providecommand{\doi}[1]{doi: #1}\else
  \providecommand{\doi}{doi: \begingroup \urlstyle{rm}\Url}\fi

\bibitem[Bai et~al.(2024)Bai, Zhou, Cemri, Pan, Suhr, Levine, and Kumar]{bai2024digirl}
Hao Bai, Yifei Zhou, Mert Cemri, Jiayi Pan, Alane Suhr, Sergey Levine, and Aviral Kumar.
\newblock Digirl: Training in-the-wild device-control agents with autonomous reinforcement learning.
\newblock \emph{arXiv preprint arXiv:2406.11896}, 2024.

\bibitem[Bai et~al.(2023)Bai, Bai, Yang, Wang, Tan, Wang, Lin, Zhou, and Zhou]{bai2023qwen}
Jinze Bai, Shuai Bai, Shusheng Yang, Shijie Wang, Sinan Tan, Peng Wang, Junyang Lin, Chang Zhou, and Jingren Zhou.
\newblock Qwen-vl: A frontier large vision-language model with versatile abilities.
\newblock \emph{arXiv preprint arXiv:2308.12966}, 2023.

\bibitem[Caesar et~al.(2018)Caesar, Uijlings, and Ferrari]{caesar2018coco}
Holger Caesar, Jasper Uijlings, and Vittorio Ferrari.
\newblock Coco-stuff: Thing and stuff classes in context.
\newblock In \emph{CVPR}, 2018.

\bibitem[Cao et~al.(2024)Cao, Gui, and Wang]{cao2024emerging}
Shengcao Cao, Liang-Yan Gui, and Yu-Xiong Wang.
\newblock Emerging pixel grounding in large multimodal models without grounding supervision.
\newblock \emph{arXiv preprint arXiv:2410.08209}, 2024.

\bibitem[Cha et~al.(2024)Cha, Kang, Mun, and Roh]{cha2024honeybee}
Junbum Cha, Wooyoung Kang, Jonghwan Mun, and Byungseok Roh.
\newblock Honeybee: Locality-enhanced projector for multimodal llm.
\newblock In \emph{Proceedings of the IEEE/CVF Conference on Computer Vision and Pattern Recognition}, pages 13817--13827, 2024.

\bibitem[Chen et~al.(2025{\natexlab{a}})Chen, Li, Zhao, Song, and Vinci]{chen2025r1v}
Liang Chen, Lei Li, Haozhe Zhao, Yifan Song, and Vinci.
\newblock R1-v:reinforcing super generalization ability in vision-language models with less than \$3.
\newblock \url{https://github.com/Deep-Agent/R1-V}, 2025{\natexlab{a}}.
\newblock Accessed: 2025-02-02.

\bibitem[Chen et~al.(2022{\natexlab{a}})Chen, Saxena, Li, Lin, Fleet, and Hinton]{chen2022unified}
Ting Chen, Saurabh Saxena, Lala Li, Tsung-Yi Lin, David~J Fleet, and Geoffrey~E Hinton.
\newblock A unified sequence interface for vision tasks.
\newblock \emph{Advances in Neural Information Processing Systems}, 35:\penalty0 31333--31346, 2022{\natexlab{a}}.

\bibitem[Chen et~al.(2024)Chen, Mees, Kumar, and Levine]{chen2024vision}
William Chen, Oier Mees, Aviral Kumar, and Sergey Levine.
\newblock Vision-language models provide promptable representations for reinforcement learning.
\newblock \emph{arXiv preprint arXiv:2402.02651}, 2024.

\bibitem[Chen et~al.(2022{\natexlab{b}})Chen, Zhao, Zhang, Duan, Qi, and Zhao]{chen2022focalclick}
Xi Chen, Zhiyan Zhao, Yilei Zhang, Manni Duan, Donglian Qi, and Hengshuang Zhao.
\newblock Focalclick: Towards practical interactive image segmentation.
\newblock In \emph{Proceedings of the IEEE/CVF Conference on Computer Vision and Pattern Recognition}, pages 1300--1309, 2022{\natexlab{b}}.

\bibitem[Chen et~al.(2025{\natexlab{b}})Chen, Li, Sun, Wang, and Chen]{chen2025sam4mllm}
Yi-Chia Chen, Wei-Hua Li, Cheng Sun, Yu-Chiang~Frank Wang, and Chu-Song Chen.
\newblock Sam4mllm: Enhance multi-modal large language model for referring expression segmentation.
\newblock In \emph{European Conference on Computer Vision}, pages 323--340. Springer, 2025{\natexlab{b}}.

\bibitem[Ding et~al.(2020)Ding, Cohen, Price, and Jiang]{ding2020phraseclick}
Henghui Ding, Scott Cohen, Brian Price, and Xudong Jiang.
\newblock Phraseclick: toward achieving flexible interactive segmentation by phrase and click.
\newblock In \emph{Computer Vision--ECCV 2020: 16th European Conference, Glasgow, UK, August 23--28, 2020, Proceedings, Part III 16}, pages 417--435. Springer, 2020.

\bibitem[Fan et~al.(2024)Fan, Zhu, Chen, Liu, Wu, Zhang, and Shen]{fan2024divergen}
Chengxiang Fan, Muzhi Zhu, Hao Chen, Yang Liu, Weijia Wu, Huaqi Zhang, and Chunhua Shen.
\newblock Divergen: Improving instance segmentation by learning wider data distribution with more diverse generative data.
\newblock In \emph{Proceedings of the IEEE/CVF Conference on Computer Vision and Pattern Recognition}, pages 3986--3995, 2024.

\bibitem[Guo et~al.(2025)Guo, Yang, Zhang, Song, Zhang, Xu, Zhu, Ma, Wang, Bi, et~al.]{guo2025deepseek}
Daya Guo, Dejian Yang, Haowei Zhang, Junxiao Song, Ruoyu Zhang, Runxin Xu, Qihao Zhu, Shirong Ma, Peiyi Wang, Xiao Bi, et~al.
\newblock Deepseek-r1: Incentivizing reasoning capability in llms via reinforcement learning.
\newblock \emph{arXiv preprint arXiv:2501.12948}, 2025.

\bibitem[Gupta et~al.(2019)Gupta, Dollar, and Girshick]{lvis}
Agrim Gupta, Piotr Dollar, and Ross Girshick.
\newblock Lvis: A dataset for large vocabulary instance segmentation.
\newblock In \emph{Proceedings of the IEEE/CVF conference on computer vision and pattern recognition}, pages 5356--5364, 2019.

\bibitem[Huang et~al.(2023)Huang, Yang, Sun, Zhang, Cao, Jiang, and Ji]{huang2023interformer}
You Huang, Hao Yang, Ke Sun, Shengchuan Zhang, Liujuan Cao, Guannan Jiang, and Rongrong Ji.
\newblock Interformer: Real-time interactive image segmentation.
\newblock In \emph{Proceedings of the IEEE/CVF International Conference on Computer Vision}, pages 22301--22311, 2023.

\bibitem[Ke et~al.(2024)Ke, Ye, Danelljan, Tai, Tang, Yu, et~al.]{ke2024segment}
Lei Ke, Mingqiao Ye, Martin Danelljan, Yu-Wing Tai, Chi-Keung Tang, Fisher Yu, et~al.
\newblock Segment anything in high quality.
\newblock \emph{Advances in Neural Information Processing Systems}, 36, 2024.

\bibitem[Kirillov et~al.(2023)Kirillov, Mintun, Ravi, Mao, Rolland, Gustafson, Xiao, Whitehead, Berg, Lo, et~al.]{sam}
Alexander Kirillov, Eric Mintun, Nikhila Ravi, Hanzi Mao, Chloe Rolland, Laura Gustafson, Tete Xiao, Spencer Whitehead, Alexander~C Berg, Wan-Yen Lo, et~al.
\newblock Segment anything.
\newblock \emph{arXiv preprint arXiv:2304.02643}, 2023.

\bibitem[Lab.(2025)]{ellm2025openr1}
EvolvingLMMs Lab.
\newblock Open r1 multimodal.
\newblock \url{https://github.com/EvolvingLMMs-Lab/open-r1-multimodal}, 2025.

\bibitem[Lai et~al.(2023)Lai, Tian, Chen, Li, Yuan, Liu, and Jia]{lisa}
Xin Lai, Zhuotao Tian, Yukang Chen, Yanwei Li, Yuhui Yuan, Shu Liu, and Jiaya Jia.
\newblock Lisa: Reasoning segmentation via large language model.
\newblock \emph{arXiv preprint arXiv:2308.00692}, 2023.

\bibitem[Lai et~al.(2024)Lai, Tian, Chen, Li, Yuan, Liu, and Jia]{lai2024lisa}
Xin Lai, Zhuotao Tian, Yukang Chen, Yanwei Li, Yuhui Yuan, Shu Liu, and Jiaya Jia.
\newblock Lisa: Reasoning segmentation via large language model.
\newblock In \emph{Proceedings of the IEEE/CVF Conference on Computer Vision and Pattern Recognition}, pages 9579--9589, 2024.

\bibitem[Lan et~al.(2024)Lan, Chen, Zhou, Xu, Ke, Wang, Feng, and Zhang]{lan2024text4segreimaginingimagesegmentation}
Mengcheng Lan, Chaofeng Chen, Yue Zhou, Jiaxing Xu, Yiping Ke, Xinjiang Wang, Litong Feng, and Wayne Zhang.
\newblock Text4seg: Reimagining image segmentation as text generation, 2024.

\bibitem[Li et~al.(2022)Li, Li, Xiong, and Hoi]{blip}
Junnan Li, Dongxu Li, Caiming Xiong, and Steven Hoi.
\newblock Blip: Bootstrapping language-image pre-training for unified vision-language understanding and generation.
\newblock In \emph{International Conference on Machine Learning}, pages 12888--12900. PMLR, 2022.

\bibitem[Li et~al.(2023)Li, Li, Savarese, and Hoi]{blip2}
Junnan Li, Dongxu Li, Silvio Savarese, and Steven Hoi.
\newblock Blip-2: Bootstrapping language-image pre-training with frozen image encoders and large language models.
\newblock \emph{arXiv preprint arXiv:2301.12597}, 2023.

\bibitem[Liew et~al.(2021)Liew, Cohen, Price, Mai, and Feng]{liew2021deep}
Jun~Hao Liew, Scott Cohen, Brian Price, Long Mai, and Jiashi Feng.
\newblock Deep interactive thin object selection.
\newblock In \emph{Proceedings of the IEEE/CVF Winter Conference on Applications of Computer Vision}, pages 305--314, 2021.

\bibitem[Lightman et~al.(2023)Lightman, Kosaraju, Burda, Edwards, Baker, Lee, Leike, Schulman, Sutskever, and Cobbe]{lightman2023let}
Hunter Lightman, Vineet Kosaraju, Yura Burda, Harri Edwards, Bowen Baker, Teddy Lee, Jan Leike, John Schulman, Ilya Sutskever, and Karl Cobbe.
\newblock Let's verify step by step.
\newblock \emph{arXiv preprint arXiv:2305.20050}, 2023.

\bibitem[Lin et~al.(2014)Lin, Maire, Belongie, Hays, Perona, Ramanan, Doll{\'a}r, and Zitnick]{lin2014microsoft}
Tsung-Yi Lin, Michael Maire, Serge Belongie, James Hays, Pietro Perona, Deva Ramanan, Piotr Doll{\'a}r, and C~Lawrence Zitnick.
\newblock Microsoft coco: Common objects in context.
\newblock In \emph{ECCV}, 2014.

\bibitem[Liu et~al.(2024{\natexlab{a}})Liu, Li, Li, and Lee]{liu2024improved}
Haotian Liu, Chunyuan Li, Yuheng Li, and Yong~Jae Lee.
\newblock Improved baselines with visual instruction tuning.
\newblock In \emph{Proceedings of the IEEE/CVF Conference on Computer Vision and Pattern Recognition}, pages 26296--26306, 2024{\natexlab{a}}.

\bibitem[Liu et~al.(2024{\natexlab{b}})Liu, Li, Li, Li, Zhang, Shen, and Lee]{liu2024llava}
Haotian Liu, Chunyuan Li, Yuheng Li, Bo Li, Yuanhan Zhang, Sheng Shen, and Yong~Jae Lee.
\newblock Llava-next: Improved reasoning, ocr, and world knowledge, 2024{\natexlab{b}}.

\bibitem[Liu et~al.(2023{\natexlab{a}})Liu, Ding, Cai, Zhang, Satzoda, Mahadevan, and Manmatha]{liu2023polyformer}
Jiang Liu, Hui Ding, Zhaowei Cai, Yuting Zhang, Ravi~Kumar Satzoda, Vijay Mahadevan, and R Manmatha.
\newblock Polyformer: Referring image segmentation as sequential polygon generation.
\newblock In \emph{CVPR}, 2023{\natexlab{a}}.

\bibitem[Liu et~al.(2023{\natexlab{b}})Liu, Xu, Bertasius, and Niethammer]{liu2023simpleclick}
Qin Liu, Zhenlin Xu, Gedas Bertasius, and Marc Niethammer.
\newblock Simpleclick: Interactive image segmentation with simple vision transformers.
\newblock In \emph{Proceedings of the IEEE/CVF International Conference on Computer Vision}, pages 22290--22300, 2023{\natexlab{b}}.

\bibitem[Liu et~al.(2023{\natexlab{c}})Liu, Zhu, Li, Chen, Wang, and Shen]{liu2023matcher}
Yang Liu, Muzhi Zhu, Hengtao Li, Hao Chen, Xinlong Wang, and Chunhua Shen.
\newblock Matcher: Segment anything with one shot using all-purpose feature matching.
\newblock \emph{arXiv preprint arXiv:2305.13310}, 2023{\natexlab{c}}.

\bibitem[Liu et~al.(2025{\natexlab{a}})Liu, Jing, Li, Zhu, Chen, Wang, and Shen]{liu2025simple}
Yang Liu, Chenchen Jing, Hengtao Li, Muzhi Zhu, Hao Chen, Xinlong Wang, and Chunhua Shen.
\newblock A simple image segmentation framework via in-context examples.
\newblock \emph{Advances in Neural Information Processing Systems}, 37:\penalty0 25095--25119, 2025{\natexlab{a}}.

\bibitem[Liu et~al.(2025{\natexlab{b}})Liu, Peng, Zhong, Yue, Lu, Yu, and Jia]{liu2025segzero}
Yuqi Liu, Bohao Peng, Zhisheng Zhong, Zihao Yue, Fanbin Lu, Bei Yu, and Jiaya Jia.
\newblock Seg-zero: Reasoning-chain guided segmentation via cognitive reinforcement.
\newblock \emph{arXiv preprint arXiv:2503.06520}, 2025{\natexlab{b}}.

\bibitem[Liu et~al.(2022)Liu, Mao, Wu, Feichtenhofer, Darrell, and Xie]{liu2022convnet}
Zhuang Liu, Hanzi Mao, Chao-Yuan Wu, Christoph Feichtenhofer, Trevor Darrell, and Saining Xie.
\newblock A convnet for the 2020s.
\newblock In \emph{Proceedings of the IEEE/CVF conference on computer vision and pattern recognition}, pages 11976--11986, 2022.

\bibitem[Loshchilov and Hutter(2016)]{loshchilov2016sgdr}
Ilya Loshchilov and Frank Hutter.
\newblock Sgdr: Stochastic gradient descent with warm restarts.
\newblock \emph{arXiv preprint arXiv:1608.03983}, 2016.

\bibitem[Loshchilov and Hutter(2017)]{adamw}
Ilya Loshchilov and Frank Hutter.
\newblock Decoupled weight decay regularization.
\newblock \emph{arXiv preprint arXiv:1711.05101}, 2017.

\bibitem[Ma et~al.(2023)Ma, Zhou, Liu, Yuan, Liu, You, and Yang]{ma2023let}
Qianli Ma, Haotian Zhou, Tingkai Liu, Jianbo Yuan, Pengfei Liu, Yang You, and Hongxia Yang.
\newblock Let's reward step by step: Step-level reward model as the navigators for reasoning.
\newblock \emph{arXiv preprint arXiv:2310.10080}, 2023.

\bibitem[Mao et~al.(2016)Mao, Huang, Toshev, Camburu, Yuille, and Murphy]{mao2016generation}
Junhua Mao, Jonathan Huang, Alexander Toshev, Oana Camburu, Alan~L Yuille, and Kevin Murphy.
\newblock Generation and comprehension of unambiguous object descriptions.
\newblock In \emph{CVPR}, 2016.

\bibitem[Mu et~al.(2023)Mu, Zhang, Hu, Wang, Ding, Jin, Wang, Dai, Qiao, and Luo]{mu2023embodiedgpt}
Yao Mu, Qinglong Zhang, Mengkang Hu, Wenhai Wang, Mingyu Ding, Jun Jin, Bin Wang, Jifeng Dai, Yu Qiao, and Ping Luo.
\newblock Embodiedgpt: Vision-language pre-training via embodied chain of thought.
\newblock \emph{arXiv preprint arXiv:2305.15021}, 2023.

\bibitem[OpenAI(2023)]{OpenAI2023GPT4}
OpenAI.
\newblock Gpt-4, 2023.

\bibitem[Ouyang et~al.(2022)Ouyang, Wu, Jiang, Almeida, Wainwright, Mishkin, Zhang, Agarwal, Slama, Ray, et~al.]{ouyang2022training}
Long Ouyang, Jeffrey Wu, Xu Jiang, Diogo Almeida, Carroll Wainwright, Pamela Mishkin, Chong Zhang, Sandhini Agarwal, Katarina Slama, Alex Ray, et~al.
\newblock Training language models to follow instructions with human feedback.
\newblock \emph{Advances in Neural Information Processing Systems}, 35:\penalty0 27730--27744, 2022.

\bibitem[Pi et~al.(2024)Pi, Yao, Gao, Zhang, and Zhang]{pi2023perceptiongpt}
Renjie Pi, Lewei Yao, Jiahui Gao, Jipeng Zhang, and Tong Zhang.
\newblock Perceptiongpt: Effectively fusing visual perception into llm.
\newblock \emph{CVPR}, 2024.

\bibitem[Pramanick et~al.(2024)Pramanick, Han, Hou, Nag, Lim, Ballas, Wang, Chellappa, and Almahairi]{pramanick2024jack}
Shraman Pramanick, Guangxing Han, Rui Hou, Sayan Nag, Ser-Nam Lim, Nicolas Ballas, Qifan Wang, Rama Chellappa, and Amjad Almahairi.
\newblock Jack of all tasks master of many: Designing general-purpose coarse-to-fine vision-language model.
\newblock In \emph{Proceedings of the IEEE/CVF Conference on Computer Vision and Pattern Recognition}, pages 14076--14088, 2024.

\bibitem[Qin et~al.(2022)Qin, Dai, Hu, Fan, Shao, and Van~Gool]{qin2022highly}
Xuebin Qin, Hang Dai, Xiaobin Hu, Deng-Ping Fan, Ling Shao, and Luc Van~Gool.
\newblock Highly accurate dichotomous image segmentation.
\newblock In \emph{European Conference on Computer Vision}, pages 38--56. Springer, 2022.

\bibitem[Rasheed et~al.(2024)Rasheed, Maaz, Shaji, Shaker, Khan, Cholakkal, Anwer, Xing, Yang, and Khan]{rasheed2024glamm}
Hanoona Rasheed, Muhammad Maaz, Sahal Shaji, Abdelrahman Shaker, Salman Khan, Hisham Cholakkal, Rao~M Anwer, Eric Xing, Ming-Hsuan Yang, and Fahad~S Khan.
\newblock Glamm: Pixel grounding large multimodal model.
\newblock In \emph{Proceedings of the IEEE/CVF Conference on Computer Vision and Pattern Recognition}, pages 13009--13018, 2024.

\bibitem[Ren et~al.(2024)Ren, Huang, Wei, Zhao, Fu, Feng, and Jin]{ren2024pixellm}
Zhongwei Ren, Zhicheng Huang, Yunchao Wei, Yao Zhao, Dongmei Fu, Jiashi Feng, and Xiaojie Jin.
\newblock Pixellm: Pixel reasoning with large multimodal model.
\newblock In \emph{Proceedings of the IEEE/CVF Conference on Computer Vision and Pattern Recognition}, pages 26374--26383, 2024.

\bibitem[Shridhar et~al.(2020)Shridhar, Thomason, Gordon, Bisk, Han, Mottaghi, Zettlemoyer, and Fox]{shridhar2020alfred}
Mohit Shridhar, Jesse Thomason, Daniel Gordon, Yonatan Bisk, Winson Han, Roozbeh Mottaghi, Luke Zettlemoyer, and Dieter Fox.
\newblock Alfred: A benchmark for interpreting grounded instructions for everyday tasks.
\newblock In \emph{Proceedings of the IEEE/CVF conference on computer vision and pattern recognition}, pages 10740--10749, 2020.

\bibitem[Siam(2025)]{siam2025pixfoundation}
Mennatullah Siam.
\newblock Pixfoundation: Are we heading in the right direction with pixel-level vision foundation models?
\newblock \emph{arXiv preprint arXiv:2502.04192}, 2025.

\bibitem[Sofiiuk et~al.(2022)Sofiiuk, Petrov, and Konushin]{sofiiuk2022reviving}
Konstantin Sofiiuk, Ilya~A Petrov, and Anton Konushin.
\newblock Reviving iterative training with mask guidance for interactive segmentation.
\newblock In \emph{2022 IEEE International Conference on Image Processing (ICIP)}, pages 3141--3145. IEEE, 2022.

\bibitem[Team et~al.(2023)Team, Anil, Borgeaud, Wu, Alayrac, Yu, Soricut, Schalkwyk, Dai, Hauth, et~al.]{Gemini}
Gemini Team, Rohan Anil, Sebastian Borgeaud, Yonghui Wu, Jean-Baptiste Alayrac, Jiahui Yu, Radu Soricut, Johan Schalkwyk, Andrew~M Dai, Anja Hauth, et~al.
\newblock Gemini: a family of highly capable multimodal models.
\newblock \emph{arXiv preprint arXiv:2312.11805}, 2023.

\bibitem[Tong et~al.(2024{\natexlab{a}})Tong, Brown, Wu, Woo, Middepogu, Akula, Yang, Yang, Iyer, Pan, et~al.]{tong2024cambrian}
Shengbang Tong, Ellis Brown, Penghao Wu, Sanghyun Woo, Manoj Middepogu, Sai~Charitha Akula, Jihan Yang, Shusheng Yang, Adithya Iyer, Xichen Pan, et~al.
\newblock Cambrian-1: A fully open, vision-centric exploration of multimodal llms.
\newblock \emph{arXiv preprint arXiv:2406.16860}, 2024{\natexlab{a}}.

\bibitem[Tong et~al.(2024{\natexlab{b}})Tong, Liu, Zhai, Ma, LeCun, and Xie]{tong2024eyes}
Shengbang Tong, Zhuang Liu, Yuexiang Zhai, Yi Ma, Yann LeCun, and Saining Xie.
\newblock Eyes wide shut? exploring the visual shortcomings of multimodal llms.
\newblock \emph{arXiv preprint arXiv:2401.06209}, 2024{\natexlab{b}}.

\bibitem[Wang et~al.(2024)Wang, Bai, Tan, Wang, Fan, Bai, Chen, Liu, Wang, Ge, et~al.]{wang2024qwen2}
Peng Wang, Shuai Bai, Sinan Tan, Shijie Wang, Zhihao Fan, Jinze Bai, Keqin Chen, Xuejing Liu, Jialin Wang, Wenbin Ge, et~al.
\newblock Qwen2-vl: Enhancing vision-language model's perception of the world at any resolution.
\newblock \emph{arXiv preprint arXiv:2409.12191}, 2024.

\bibitem[Wang et~al.(2022)Wang, Lu, Li, Tao, Guo, Gong, and Liu]{wang2022cris}
Zhaoqing Wang, Yu Lu, Qiang Li, Xunqiang Tao, Yandong Guo, Mingming Gong, and Tongliang Liu.
\newblock Cris: Clip-driven referring image segmentation.
\newblock In \emph{CVPR}, 2022.

\bibitem[Wei et~al.(2022)Wei, Wang, Schuurmans, Bosma, Xia, Chi, Le, Zhou, et~al.]{wei2022chain}
Jason Wei, Xuezhi Wang, Dale Schuurmans, Maarten Bosma, Fei Xia, Ed Chi, Quoc~V Le, Denny Zhou, et~al.
\newblock Chain-of-thought prompting elicits reasoning in large language models.
\newblock \emph{Advances in Neural Information Processing Systems}, 35:\penalty0 24824--24837, 2022.

\bibitem[Xia et~al.(2024)Xia, Han, Han, Pan, Song, and Huang]{xia2023gsva}
Zhuofan Xia, Dongchen Han, Yizeng Han, Xuran Pan, Shiji Song, and Gao Huang.
\newblock Gsva: Generalized segmentation via multimodal large language models.
\newblock \emph{CVPR}, 2024.

\bibitem[Yang et~al.(2023)Yang, Qu, Lai, Tian, Peng, Liu, and Jia]{lisa++}
Senqiao Yang, Tianyuan Qu, Xin Lai, Zhuotao Tian, Bohao Peng, Shu Liu, and Jiaya Jia.
\newblock An improved baseline for reasoning segmentation with large language model.
\newblock \emph{arXiv preprint arXiv:2312.17240}, 2023.

\bibitem[Yang et~al.(2025)Yang, He, Pan, Jiang, Deng, Yang, Lu, Zhu, and Zhang]{yang2025r1one}
Yi Yang, Xiaoxuan He, Hongkun Pan, Xiyan Jiang, Yan Deng, Xingtao Yang, Haoyu Lu, Minfeng Zhu, and Bo Zhang.
\newblock R1-onevision:open-source multimodal large language model with reasoning ability.
\newblock \url{https://github.com/Fancy-MLLM/R1-Onevision}, 2025.
\newblock Accessed: 2025-02-02.

\bibitem[Yang et~al.(2022)Yang, Wang, Tang, Chen, Zhao, and Torr]{yang2022lavt}
Zhao Yang, Jiaqi Wang, Yansong Tang, Kai Chen, Hengshuang Zhao, and Philip~HS Torr.
\newblock Lavt: Language-aware vision transformer for referring image segmentation.
\newblock In \emph{CVPR}, 2022.

\bibitem[Yao et~al.(2024{\natexlab{a}})Yao, Li, Ren, Wang, Liu, Sun, and Hou]{yao2024deco}
Linli Yao, Lei Li, Shuhuai Ren, Lean Wang, Yuanxin Liu, Xu Sun, and Lu Hou.
\newblock Deco: Decoupling token compression from semantic abstraction in multimodal large language models.
\newblock \emph{arXiv preprint arXiv:2405.20985}, 2024{\natexlab{a}}.

\bibitem[Yao et~al.(2023{\natexlab{a}})Yao, Yu, Zhao, Shafran, Griffiths, Cao, and Narasimhan]{yao2023tree}
Shunyu Yao, Dian Yu, Jeffrey Zhao, Izhak Shafran, Thomas~L Griffiths, Yuan Cao, and Karthik Narasimhan.
\newblock Tree of thoughts: Deliberate problem solving with large language models.
\newblock \emph{arXiv preprint arXiv:2305.10601}, 2023{\natexlab{a}}.

\bibitem[Yao et~al.(2023{\natexlab{b}})Yao, Zhao, Yu, Du, Shafran, Narasimhan, and Cao]{yao2023react}
Shunyu Yao, Jeffrey Zhao, Dian Yu, Nan Du, Izhak Shafran, Karthik~R Narasimhan, and Yuan Cao.
\newblock React: Synergizing reasoning and acting in language models.
\newblock In \emph{The Eleventh International Conference on Learning Representations}, 2023{\natexlab{b}}.

\bibitem[Yao et~al.(2024{\natexlab{b}})Yao, Yu, Zhao, Shafran, Griffiths, Cao, and Narasimhan]{yao2024tree}
Shunyu Yao, Dian Yu, Jeffrey Zhao, Izhak Shafran, Tom Griffiths, Yuan Cao, and Karthik Narasimhan.
\newblock Tree of thoughts: Deliberate problem solving with large language models.
\newblock \emph{Advances in Neural Information Processing Systems}, 36, 2024{\natexlab{b}}.

\bibitem[Ying et~al.(2024)Ying, Meng, Wang, Li, Lin, Yang, Zhang, Zhang, Lin, Liu, et~al.]{ying2024mmt}
Kaining Ying, Fanqing Meng, Jin Wang, Zhiqian Li, Han Lin, Yue Yang, Hao Zhang, Wenbo Zhang, Yuqi Lin, Shuo Liu, et~al.
\newblock Mmt-bench: A comprehensive multimodal benchmark for evaluating large vision-language models towards multitask agi.
\newblock \emph{arXiv preprint arXiv:2404.16006}, 2024.

\bibitem[Yu et~al.(2016{\natexlab{a}})Yu, Poirson, Yang, Berg, and Berg]{refcoco}
Licheng Yu, Patrick Poirson, Shan Yang, Alexander~C Berg, and Tamara~L Berg.
\newblock Modeling context in referring expressions.
\newblock In \emph{Computer Vision--ECCV 2016: 14th European Conference, Amsterdam, The Netherlands, October 11-14, 2016, Proceedings, Part II 14}, pages 69--85. Springer, 2016{\natexlab{a}}.

\bibitem[Yu et~al.(2016{\natexlab{b}})Yu, Poirson, Yang, Berg, and Berg]{yu2016modeling}
Licheng Yu, Patrick Poirson, Shan Yang, Alexander~C Berg, and Tamara~L Berg.
\newblock Modeling context in referring expressions.
\newblock In \emph{ECCV}, 2016{\natexlab{b}}.

\bibitem[Yu et~al.(2018)Yu, Lin, Shen, Yang, Lu, Bansal, and Berg]{yu2018mattnet}
Licheng Yu, Zhe Lin, Xiaohui Shen, Jimei Yang, Xin Lu, Mohit Bansal, and Tamara~L Berg.
\newblock Mattnet: Modular attention network for referring expression comprehension.
\newblock In \emph{CVPR}, 2018.

\bibitem[Yuan et~al.(2024{\natexlab{a}})Yuan, Yuan, Tan, Wang, Huang, and Huang]{yuan2024rrhf}
Hongyi Yuan, Zheng Yuan, Chuanqi Tan, Wei Wang, Songfang Huang, and Fei Huang.
\newblock Rrhf: Rank responses to align language models with human feedback.
\newblock \emph{Advances in Neural Information Processing Systems}, 36, 2024{\natexlab{a}}.

\bibitem[Yuan et~al.(2024{\natexlab{b}})Yuan, Li, Liu, Tang, Luo, Qin, Zhang, and Zhu]{yuan2024osprey}
Yuqian Yuan, Wentong Li, Jian Liu, Dongqi Tang, Xinjie Luo, Chi Qin, Lei Zhang, and Jianke Zhu.
\newblock Osprey: Pixel understanding with visual instruction tuning.
\newblock In \emph{Proceedings of the IEEE/CVF Conference on Computer Vision and Pattern Recognition}, pages 28202--28211, 2024{\natexlab{b}}.

\bibitem[Zelikman et~al.(2022)Zelikman, Wu, Mu, and Goodman]{zelikman2022star}
Eric Zelikman, Yuhuai Wu, Jesse Mu, and Noah Goodman.
\newblock Star: Bootstrapping reasoning with reasoning.
\newblock \emph{Advances in Neural Information Processing Systems}, 35:\penalty0 15476--15488, 2022.

\bibitem[Zhai et~al.(2024)Zhai, Bai, Lin, Pan, Tong, Zhou, Suhr, Xie, LeCun, Ma, et~al.]{zhai2024fine}
Yuexiang Zhai, Hao Bai, Zipeng Lin, Jiayi Pan, Shengbang Tong, Yifei Zhou, Alane Suhr, Saining Xie, Yann LeCun, Yi Ma, et~al.
\newblock Fine-tuning large vision-language models as decision-making agents via reinforcement learning.
\newblock \emph{arXiv preprint arXiv:2405.10292}, 2024.

\bibitem[Zhang et~al.(2024{\natexlab{a}})Zhang, Wu, Teng, Liao, Xu, Xiao, Wei, and Tang]{zhang2024android}
Jiwen Zhang, Jihao Wu, Yihua Teng, Minghui Liao, Nuo Xu, Xiao Xiao, Zhongyu Wei, and Duyu Tang.
\newblock Android in the zoo: Chain-of-action-thought for gui agents.
\newblock \emph{arXiv preprint arXiv:2403.02713}, 2024{\natexlab{a}}.

\bibitem[Zhang et~al.(2024{\natexlab{b}})Zhang, Li, Fei, Yuan, Wu, Ji, Loy, and Yan]{zhang2024omg}
Tao Zhang, Xiangtai Li, Hao Fei, Haobo Yuan, Shengqiong Wu, Shunping Ji, Chen~Change Loy, and Shuicheng Yan.
\newblock Omg-llava: Bridging image-level, object-level, pixel-level reasoning and understanding.
\newblock \emph{arXiv preprint arXiv:2406.19389}, 2024{\natexlab{b}}.

\bibitem[Zhang and Zhang(2023)]{zhang2023you}
Zhuosheng Zhang and Aston Zhang.
\newblock You only look at screens: Multimodal chain-of-action agents.
\newblock \emph{arXiv preprint arXiv:2309.11436}, 2023.

\bibitem[Zhang et~al.(2024{\natexlab{c}})Zhang, Ma, Zhang, and Bai]{zhang2025psalm}
Zheng Zhang, Yeyao Ma, Enming Zhang, and Xiang Bai.
\newblock Psalm: Pixelwise segmentation with large multi-modal model.
\newblock In \emph{European Conference on Computer Vision}, pages 74--91, 2024{\natexlab{c}}.

\bibitem[Zhao et~al.(2025)Zhao, Liu, Zheng, Zhu, Zhao, Chen, He, and Shen]{zhao2025diception}
Canyu Zhao, Mingyu Liu, Huanyi Zheng, Muzhi Zhu, Zhiyue Zhao, Hao Chen, Tong He, and Chunhua Shen.
\newblock Diception: A generalist diffusion model for visual perceptual tasks.
\newblock \emph{arXiv preprint arXiv:2502.17157}, 2025.

\bibitem[Zhu et~al.(2023{\natexlab{a}})Zhu, Sharma, Frujeri, Dong, Zhu, Jordan, and Jiao]{zhu2023fine}
Banghua Zhu, Hiteshi Sharma, Felipe~Vieira Frujeri, Shi Dong, Chenguang Zhu, Michael~I Jordan, and Jiantao Jiao.
\newblock Fine-tuning language models with advantage-induced policy alignment.
\newblock \emph{arXiv preprint arXiv:2306.02231}, 2023{\natexlab{a}}.

\bibitem[Zhu et~al.(2023{\natexlab{b}})Zhu, Li, Chen, Fan, Mao, Jing, Liu, and Shen]{zhu2023segprompt}
Muzhi Zhu, Hengtao Li, Hao Chen, Chengxiang Fan, Weian Mao, Chenchen Jing, Yifan Liu, and Chunhua Shen.
\newblock Segprompt: Boosting open-world segmentation via category-level prompt learning.
\newblock In \emph{Proceedings of the IEEE/CVF International Conference on Computer Vision}, pages 999--1008, 2023{\natexlab{b}}.

\bibitem[Zhu et~al.(2024{\natexlab{a}})Zhu, Fan, Chen, Liu, Mao, Xu, and Shen]{zhu2024generative}
Muzhi Zhu, Chengxiang Fan, Hao Chen, Yang Liu, Weian Mao, Xiaogang Xu, and Chunhua Shen.
\newblock Generative active learning for long-tailed instance segmentation.
\newblock \emph{arXiv preprint arXiv:2406.02435}, 2024{\natexlab{a}}.

\bibitem[Zhu et~al.(2024{\natexlab{b}})Zhu, Liu, Luo, Jing, Chen, Xu, Wang, and Shen]{zhu2024unleashing}
Muzhi Zhu, Yang Liu, Zekai Luo, Chenchen Jing, Hao Chen, Guangkai Xu, Xinlong Wang, and Chunhua Shen.
\newblock Unleashing the potential of the diffusion model in few-shot semantic segmentation.
\newblock \emph{arXiv preprint arXiv:2410.02369}, 2024{\natexlab{b}}.

\bibitem[Zou et~al.(2023)Zou, Dou, Yang, Gan, Li, Li, Dai, Behl, Wang, Yuan, et~al.]{zou2023generalized}
Xueyan Zou, Zi-Yi Dou, Jianwei Yang, Zhe Gan, Linjie Li, Chunyuan Li, Xiyang Dai, Harkirat Behl, Jianfeng Wang, Lu Yuan, et~al.
\newblock Generalized decoding for pixel, image, and language.
\newblock In \emph{CVPR}, 2023.

\bibitem[Zou et~al.(2024)Zou, Yang, Zhang, Li, Li, Wang, Wang, Gao, and Lee]{zou2024segment}
Xueyan Zou, Jianwei Yang, Hao Zhang, Feng Li, Linjie Li, Jianfeng Wang, Lijuan Wang, Jianfeng Gao, and Yong~Jae Lee.
\newblock Segment everything everywhere all at once.
\newblock \emph{NeurIPS}, 2024.

\end{thebibliography}
